\documentclass[sigconf, nonacm, screen]{acmart}
\AtBeginDocument{%
  }

\setcopyright{acmlicensed}
\copyrightyear{2026}
\acmYear{2026}
\acmDOI{XXXXXXX.XXXXXXX}
\acmConference[Web'2026]{Web}{April 13--17,
  2026}{Dubai, United Arab Emirates}
\acmISBN{978-1-4503-XXXX-X/2018/06}

\usepackage{amsmath,amsfonts}
\usepackage{algorithmic}
\usepackage{graphicx}
\usepackage{enumerate}
\usepackage{bm}

\usepackage{textcomp}
\usepackage{xcolor}
\usepackage{enumitem}

\usepackage{url}
\usepackage{caption}
\usepackage{float}

\usepackage{balance}
\usepackage{amsthm}
\usepackage{multirow}

\theoremstyle{plain}

\newtheorem*{probs}{Problem Statement}
\newtheoremstyle{myitalic}            
{3pt}   
{3pt}   
{} 
{}      
{\bfseries} 
{.}     
{ }     
{}      
\theoremstyle{myitalic}  
\newtheorem{definition}{Definition}

\makeatletter
\newcommand{\Hline}{%
  \noalign{\ifnum0=`}\fi\hrule height 1.2pt
  \futurelet\reserved@a\@xhline}
\makeatother
\begin{document}

\author{Soheila Farokhi}
\email{soheila.farokhi@usu.edu}
\affiliation{%
  \institution{Utah State University}
  \city{Logan}
  \state{Utah}
  \country{USA}
}

\author{Xiaojun Qi}
\email{xiaojun.qi@usu.edu}
\affiliation{%
  \institution{Utah State University}
  \city{Logan}
  \state{Utah}
  \country{USA}
}

\author{Hamid Karimi}
\email{hamid.karimi@usu.edu}
\affiliation{%
  \institution{Utah State University}
  \city{Logan}
  \state{Utah}
  \country{USA}
}

\renewcommand{\shortauthors}{Farokhi et al.}

\title{TAWRMAC: A Novel Dynamic Graph Representation Learning Method}

\begin{abstract}
Dynamic graph representation learning has become essential for analyzing evolving networks in domains such as social network analysis, recommendation systems, and traffic analysis. However, existing continuous-time methods face three key challenges: (1) some methods depend solely on node-specific memory without effectively incorporating information from neighboring nodes, resulting in embedding \textit{staleness}; (2) most fail to explicitly capture correlations between node neighborhoods, limiting \textit{contextual awareness}; and (3) many fail to fully capture the \textit{structural dynamics} of evolving graphs, especially in absence of rich link attributes. To address these limitations, we introduce \mbox{\textbf{TAWRMAC}}—a novel framework that integrates \textbf{\underline{T}}emporal \textbf{\underline{A}}nonymous \textbf{\underline{W}}alks with \textbf{\underline{R}}estart, \textbf{\underline{M}}emory \textbf{\underline{A}}ugmentation, and Neighbor \textbf{\underline{C}}o-occurrence embedding. TAWRMAC enhances embedding stability through a memory-augmented GNN with fixed-time encoding and improves contextual representation by explicitly capturing neighbor correlations. Additionally, its Temporal Anonymous Walks with Restart mechanism distinguishes between nodes exhibiting repetitive interactions and those forming new connections beyond their immediate neighborhood. This approach captures structural dynamics better and supports strong inductive learning. Extensive experiments on multiple benchmark datasets demonstrate that TAWRMAC consistently outperforms state-of-the-art methods in dynamic link prediction and node classification under both \textit{transductive} and \textit{inductive} settings across three different negative sampling strategies. By providing stable, generalizable, and context-aware embeddings, TAWRMAC advances the state of the art in continuous-time dynamic graph learning. The code is  available at~\url{https://anonymous.4open.science/r/tawrmac-A253}.
\end{abstract}  



\maketitle

 \section{Introduction}
\label{sec:introduction}


Dynamic graphs offer a flexible framework to model the temporal evolution of interactions between entities, making them particularly well-suited for domains such as social networks~\cite{sun2022ddgcn,kheiri2023analysis,farokhi2024edge,cheng2025bts}, education~\cite{karimi2020online}, recommendation systems~\cite{ li2020dynamic}, and traffic analysis~\cite{jiang2022graph}. By integrating time-dependent information, dynamic graphs provide a richer context compared to static graph representations, which often overlook the sequential nature of interactions. This capability has driven growing interest in dynamic graph representation learning, leading to notable progress in understanding and predicting complex systems~\cite{kazemi2020representation, skarding2021foundations, barros2021survey}. Continuous-time modeling approaches~\cite{kumar2019predicting, trivedi2019dyrep, xu2020inductive, rossi2020temporal,ma2020streaming, wang2021inductive, wang2021tcl,luo2022neighborhood, cong2023we,yu2023towards} have gained particular attention for their ability to effectively capture the temporal dynamics of interactions, offering enhanced adaptability and precision compared to discrete-time methods~\cite{goyal2018dyngem, goyal2020dyngraph2vec, sankar2020dysat, pareja2020evolvegcn,you2022roland}.

\begin{figure}[htb]
  \vspace{-3mm}
    \centering
    \includegraphics[width=0.88\columnwidth]{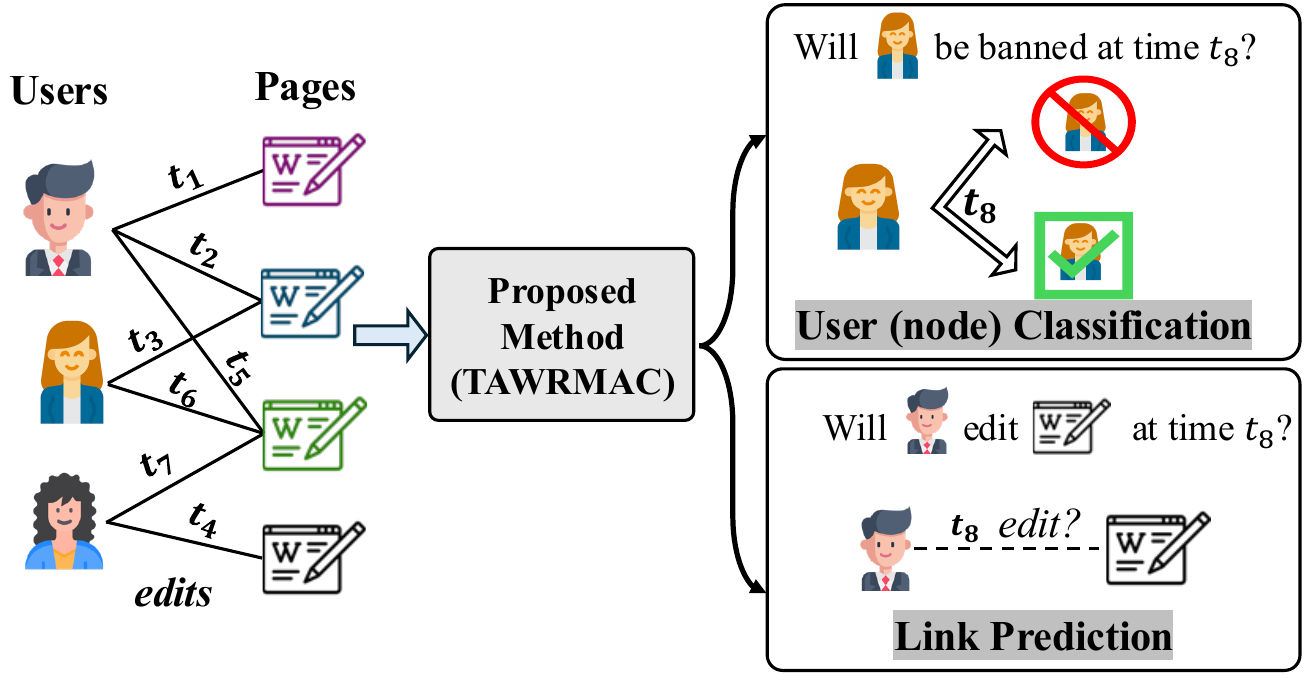}
    \vspace{-6pt}
    \caption{
    An example of a dynamic graph showing user interactions with Wikipedia pages over time. Nodes represent users and pages, while edges indicate interactions at specific timestamps. A dynamic graph representation learning model, such as TAWRMAC, captures the temporal evolution of these interactions to enable downstream tasks such as predicting future edits or user bans.
}
    \label{fig:motivation}
    \vspace{-3mm}
\end{figure}

Figure~\ref{fig:motivation} illustrates an example of a dynamic graph representing users editing Wikipedia pages over time. In this graph, nodes represent users and pages, while edges are interactions between them, such as page edits, along with their associated timestamps. A dynamic graph representation learning model like ours leverages the temporal and structural information in such graphs to learn latent representations of entities (nodes) and interactions. These representations enable key downstream tasks, such as predicting whether a user will edit a specific Wikipedia page at a given time or determining whether a user might face a ban in the future.

Despite significant progress in continuous-time dynamic graph representation learning, existing methods face several key limitations. \textbf{First}, the use of node-specific memory has shown promise in modeling long-term dependencies, as demonstrated by several successful prior architectures~\cite{kumar2019predicting, rossi2020temporal,trivedi2019dyrep}. However, these approaches often lack neighbor-based aggregation such as graph neural networks (GNNs), leaving them vulnerable to embedding \textit{staleness}, where outdated representations fail to capture recent graph dynamics~\cite{kumar2019predicting, trivedi2019dyrep}. Additionally, some of these approaches rely solely on trainable time encoding functions~\cite{rossi2020temporal, kumar2019predicting}, which can introduce instability during training due to issues like exploding gradients. In contrast, fixed time encodings provide greater stability and robustness, mitigating these challenges~\cite{cong2023we}.

\textbf{Secondly}, many models compute node embeddings for each interaction independently, without considering the interdependence of the nodes’ temporal neighborhoods~\cite{xu2020inductive, rossi2020temporal, cong2023we}. This oversight neglects potentially valuable \emph{contextual information} that could enhance predictions. For example, in a temporal co-authorship network, the shared collaboration history of two nodes (authors) with a common neighbor can signal a potential future collaboration between those nodes. Incorporating mutual influences between temporal neighborhoods could yield more informative and predictive dynamic representations.


\textbf{Thirdly}, many prior methods fail to adequately capture the structural dynamics of evolving graphs. For instance, Wang et al.\cite{wang2021inductive} showed that models such as TGAT\cite{xu2020inductive} and TGN~\cite{rossi2020temporal} rely heavily on rich link attributes (e.g., textual content of a page edit in Wikipedia) and struggle to generalize when such attributes are limited. In contrast, walk-based models like CAWN~\cite{wang2021inductive} offer a more flexible and generalizable solution by modeling the evolving structure directly through anonymous walks. However, CAWN uses fixed-length walks without any \emph{restart mechanism}, limiting its ability to adaptively distinguish between repetitive local activity and exploratory node behavior. This restricts its capacity to dynamically control the balance between \textit{exploiting familiar structures} and \textit{exploring new regions of the graph}.
 
To overcome these challenges, we propose a novel dynamic graph representation learning framework featuring \textbf{\underline{T}}emporal \textbf{\underline{A}}nonymous \textbf{\underline{W}}alks with \textbf{\underline{R}}estart, \textbf{\underline{M}}emory \textbf{\underline{A}}ugmentation, and Neighbor \textbf{\underline{C}}o-occurrence embedding, called \textbf{TAWRMAC}, which introduces several key innovations to advance the state of the art. The main contributions of this paper are as follows:

\begin{itemize}

    \item \textbf{Memory-Augmented Embedding (MAE)}: To tackle the first challenge mentioned above, TAWRMAC integrates node-specific memory with a GNN-based architecture and introduces a fixed-time encoding mechanism to enhance the stability and reliability of the embeddings.
    \item \textbf{Neighbor Co-occurrence Embedding (NCE)}: TAWRMAC addresses the second challenge by capturing contextual information through modeling correlations between node neighborhoods, resulting in richer and more informative representations of node interactions.

    \item \textbf{Temporal Anonymous Walks with Restart (TAWR)}: To better capture the structural dynamics of evolving graphs, and address the third challenge, our proposed TAWR builds on anonymous walk-based approaches and introduces a novel, learnable restart mechanism. Unlike prior models such as CAWN~\cite{wang2021inductive} that use fixed-length walks without restart, TAWR enables each node to adapt its walk behavior through a time-dependent, node-specific restart probability. This allows the model to dynamically balance between local exploitation and broader structural exploration, based on the node’s interaction patterns.


    \item \textbf{Modular Integration for Efficiency}: Beyond the design of each individual component, TAWRMAC’s modular architecture allows these innovations to synergistically work together efficiently. The complementary strengths of TAWR, MAE, and NCE enable strong predictive performance without requiring computationally expensive architectures. Our extensive experiments on dynamic link prediction—across both transductive and inductive settings—and dynamic node classification tasks demonstrate TAWRMAC’s superiority over recent baselines in both performance and runtime efficiency.
\end{itemize}


\section{Related Work}
\label{sec:related_work}
\begin{figure*}[htb]
    \centering
    \includegraphics[width=0.9\textwidth]{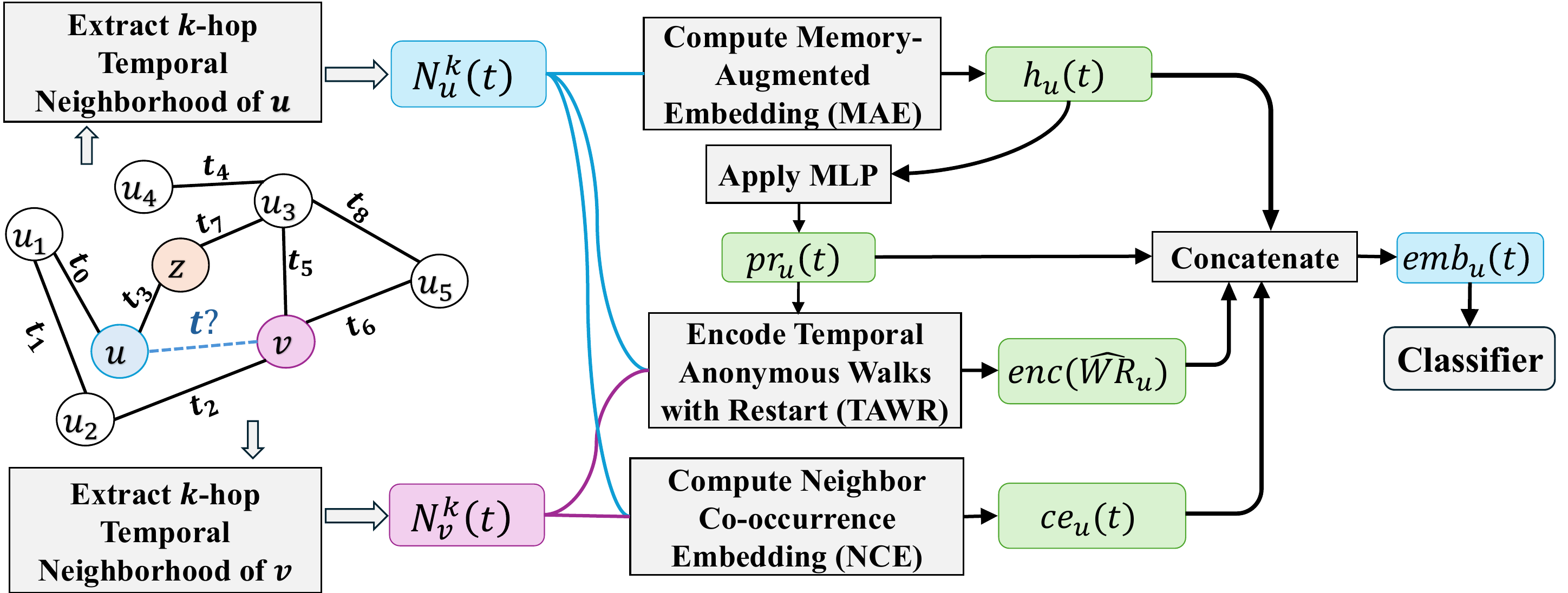}
    
    \caption{Overview of the proposed method (TAWRMAC), illustrating the computation of the embedding for node 
$u$ at time 
$t$.}
    \label{fig:model}
    \vspace{-3mm}
\end{figure*}
Dynamic graph learning methods are primarily divided into two categories: discrete-time and continuous-time models, based on how they handle evolving graph data~\cite{dynamic_graphs_survey}.

 In this work, we focus on continuous-time approaches, as they offer a more flexible and effective framework for modeling dynamic graphs. These models operate directly on evolving graphs without discretizing time, instead representing interactions as streams of events. JODIE~\cite{kumar2019predicting} models temporal bipartite networks by using two interconnected RNNs to update user and item embeddings with each interaction.
Similarly, DyRep~\cite{trivedi2019dyrep} uses RNNs to update node representations each time a new edge appears. Different from JODIE, however, DyRep  extends beyond direct interactions by considering additional neighbor hops, offering a more expressive model at the expense of increased complexity.
TGAT~\cite{xu2020inductive} uses a self-attention mechanism combined with a time encoding function to aggregate features from a nodes' temporal neighbors, similar to static GNNs like GraphSAGE~\cite{hamilton2017inductive}. However, TGAT requires storing all historical neighbors, which limits its scalability.
TGN~\cite{rossi2020temporal} extends TGAT by incorporating an evolving memory module that updates each node's state upon interaction, enhancing the model's ability to capture long-term dependencies using only limited historical neighbors and few graph attention layers.
DyGNN~\cite{ma2020streaming} includes an update component to refresh node information based on new edges and time intervals, and a propagation component to distribute interaction information to affected nodes.
CAWN~\cite{wang2021inductive} represents dynamic networks by extracting multiple causal anonymous walks for each node, replacing node identities with relative identifiers based on walk patterns. It encodes these walks using RNNs and aggregates them into a single representation using a self-attention mechanism for downstream tasks.
TCL~\cite{wang2021tcl} generates interaction sequences for each node using a breadth-first search on temporal subgraphs and employs a Transformer~\cite{vaswani2017attention} to learn node representations by integrating graph topology and temporal information. It utilizes a cross-attention mechanism to model dependencies between interacting nodes and applies contrastive learning to enhance the model’s performance.
NAT~\cite{luo2022neighborhood} maintains node and link representations, utilizing a dictionary-style neighborhood cache (N-cache) that stores historical neighbor information. It updates and computes neighborhood features by leveraging GPU parallelism and hash-based searches, enabling fast link predictions. EdgeBank~\cite{poursafaei2022towards} is a memory-based approach for dynamic link prediction, storing previously observed interactions and predicting them as positive during testing if they remain in memory.
GraphMixer~\cite{cong2023we} uses a fixed time encoding function within its MLP-based link encoder to extract information from temporal links, as it performs better than a trainable version. It also features a node encoder employing neighbor mean-pooling to summarize node information.
DyGFormer~\cite{yu2023towards} is a Transformer-based architecture for dynamic graph learning that focuses on sequences of nodes' historical first-hop interactions. It utilizes a neighbor co-occurrence encoding scheme to capture the frequency and correlation of neighbors for both source and destination nodes. However, DyGFormer ignores high-order relationships between nodes and focuses only first-hop neighbors, which may be suboptimal~\cite{yu2023towards}. FreeDyG~\cite{tian2024freedyg} builds on DyGFormer by introducing a frequency-domain perspective to dynamic graph modeling. It proposes a node interaction frequency encoding and a frequency-enhanced MLP-Mixer layer to capture periodic patterns and shifts in temporal behaviors. While this approach improves temporal pattern recognition, FreeDyG—like DyGFormer—relies solely on first-hop neighbor co-occurrence and thus fails to capture higher-order structural dependencies.

TAWRMAC belongs to the family of continuous-time dynamic graph learning methods, designed to capture both structural and temporal dynamics. Unlike prior approaches, it achieves a robust balance between \textbf{stability}, \textbf{higher-order dependency modeling}, and \textbf{inductive generalization}, leading to superior performance across diverse datasets as shown in Section~\ref{sec:results}.

\section{Preliminaries}
\label{sec:preliminaries}
\setlength{\abovedisplayskip}{5pt}
\setlength{\belowdisplayskip}{5pt}
\setlength{\abovedisplayshortskip}{0pt}
\setlength{\belowdisplayshortskip}{0pt}

\begin{definition}
A \textit{temporal graph} \( G = (V, \mathcal{E}) \) is defined by a set of nodes \( V \) and a set of temporal edges \( \mathcal{E} = \{e_{uv}(t) \mid (u, v, t) \text{ s.t. } u, v \in V \text{ and } t \in \mathbb{R}^+ \} \). Additionally, each node \( u \in V \) can be associated with a feature vector \( x_u(t) \in \mathbb{R}^d \) at any time \( t \). The feature vectors for all nodes at time \( t \) are represented as a matrix \( X(t) \in \mathbb{R}^{|V| \times d} \), where \( d \) is the dimensionality of the features.

\end{definition}

\begin{definition}
    The \textit{temporal neighborhood} of a node \( u \) in a temporal graph at time \( t \) is defined as:
    \[
    N_u(t) = \{ z \mid (u, z, t') \in \mathcal{E} \lor (z, u, t') \in \mathcal{E}, \, t' < t \},
    \]
    representing nodes directly connected to \( u \) via edges occurring before time \( t \). The \textit{k-hop temporal neighborhood} of a node \( u \) at time \( t \), denoted as \( N_u^k(t) \) (where \( k > 1 \)), consists of all nodes that can be reached from \( u \) within \( k \)-hops via temporal edges with timestamps \( t' < t \).
\end{definition}

\begin{definition}
We define a \textit{temporal walk} $W_{u,v}$ from node $u$ to node $v$ on a temporal graph as a sequence of edges:
\[W_{u,v} = [(u, u_1, t_0), (u_1, u_2, t_1), \cdots, (u_{m}, v, t_m)]
\]
where $t_{i+1}<t_i$ for $i=0,1,\cdots,m$.

\end{definition}

\begin{probs}
\normalfont Given a temporal network and a sequence of historical interactions before time \( t \), \( \{(u', v', t') \mid t' < t\} \), the goal of dynamic graph representation learning is to compute time-aware embeddings \( emb_u(t) \in \mathbb{R}^d \) and \( emb_v(t) \in \mathbb{R}^d \) for nodes \( u \) and \( v \), where \( d \) is the dimensionality of the embedding space. These embeddings should capture both the temporal and structural dynamics of the network.

The effectiveness of these embeddings is evaluated through the following tasks (see Figure~\ref{fig:motivation} for an illustration):
\begin{itemize}
    \item \textbf{Dynamic Link Prediction:} Predicting whether a link will exist between nodes \( u \) and \( v \) at time \( t \), given their historical interactions.
    \item \textbf{Dynamic Node Classification:} Inferring the state or properties of nodes \( u \) or \( v \) at time \( t \).
\end{itemize}

\end{probs}








\section{Proposed Method (TAWRMAC)}
\label{sec:methodology}

In this section, we introduce the core components of our proposed model for learning dynamic graph representations. The model (illustrated in Figure~\ref{fig:model}) includes three key components, each designed to capture a unique aspect of the temporal and structural dynamics of the graph.  First, the \textit{Memory-Augmented Embedding (MAE)} component maintains and updates node-specific memory vectors to encode historical interactions. Second, the \textit{Neighbor Co-occurrence Embedding (NCE)} component captures patterns of shared neighbors across temporal interactions to enhance structural context. Finally, the \textit{Temporal Anonymous Walks with Restart (TAWR)} component explores temporal network motifs and models node exploration and exploitation tendencies through a probabilistic restart mechanism. Together, these components enable a comprehensive representation of nodes in evolving temporal networks.

\subsection{Memory-Augmented Embedding (MAE)}
\label{subsec:mae}

We assign a memory vector \( m_u(t) \in \mathbb{R}^{d_m} \) to each node \( u \) in the graph, consistent with prior memory-based temporal GNNs \cite{rossi2020temporal}. This vector is initialized to zero the first time the node is encountered. The memory of a node is updated whenever the node participates in an event, such as the appearance of a new edge or an update to its node features. If a new edge \( e_{uv}(t) \) appears between nodes \( u \) and \( v \), the memory of both nodes involved in the event is updated as follows:
\begin{align}
    \text{msg}_u(t) &= [m_u(t^-) \Vert m_v(t^-) \Vert \phi_1(\Delta t) \Vert e_{uv}(t)],\nonumber \\
    \text{msg}_v(t) &= [m_v(t^-) \Vert m_u(t^-) \Vert \phi_1(\Delta t) \Vert e_{uv}(t)], \nonumber\\
    m_u(t) &= \text{RNN}(\text{msg}_u(t), m_u(t^-)), \nonumber\\
    m_v(t) &= \text{RNN}(\text{msg}_v(t), m_v(t^-)),\nonumber
\end{align}
where \( m_u(t^-) \) and \( m_v(t^-) \) are the memory vectors of nodes \( u \) and \( v \) before time \( t \), $\text{msg}_u(t)$ and $\text{msg}_v(t)$ are the computed messages for nodes \( u \) and \( v \), \( \phi_1(\Delta t) \) is the time encoding function defined by Xu et al.~\cite{xu2020inductive}, and RNN is a learnable recurrent neural network which can be a GRU~\cite{cho2014learning} or an LSTM~\cite{hochreiter1997long} unit.

If node \( u \) is involved in an update event where its node features are modified, the memory is updated as follows:
\begin{align}
    \text{msg}_u(t) &= [m_u(t^-) \Vert \phi_1(\Delta t) \Vert x_u(t)],\nonumber \\
    m_u(t) &= \text{RNN}(\text{msg}_u(t), m_u(t^-)),\nonumber
\end{align}
where \( x_u(t) \) is the feature vector of node \( u \) at time \( t \), and all other terms are as defined above.

\begin{figure}[htb]
    \centering
    \includegraphics[width=0.8\columnwidth]{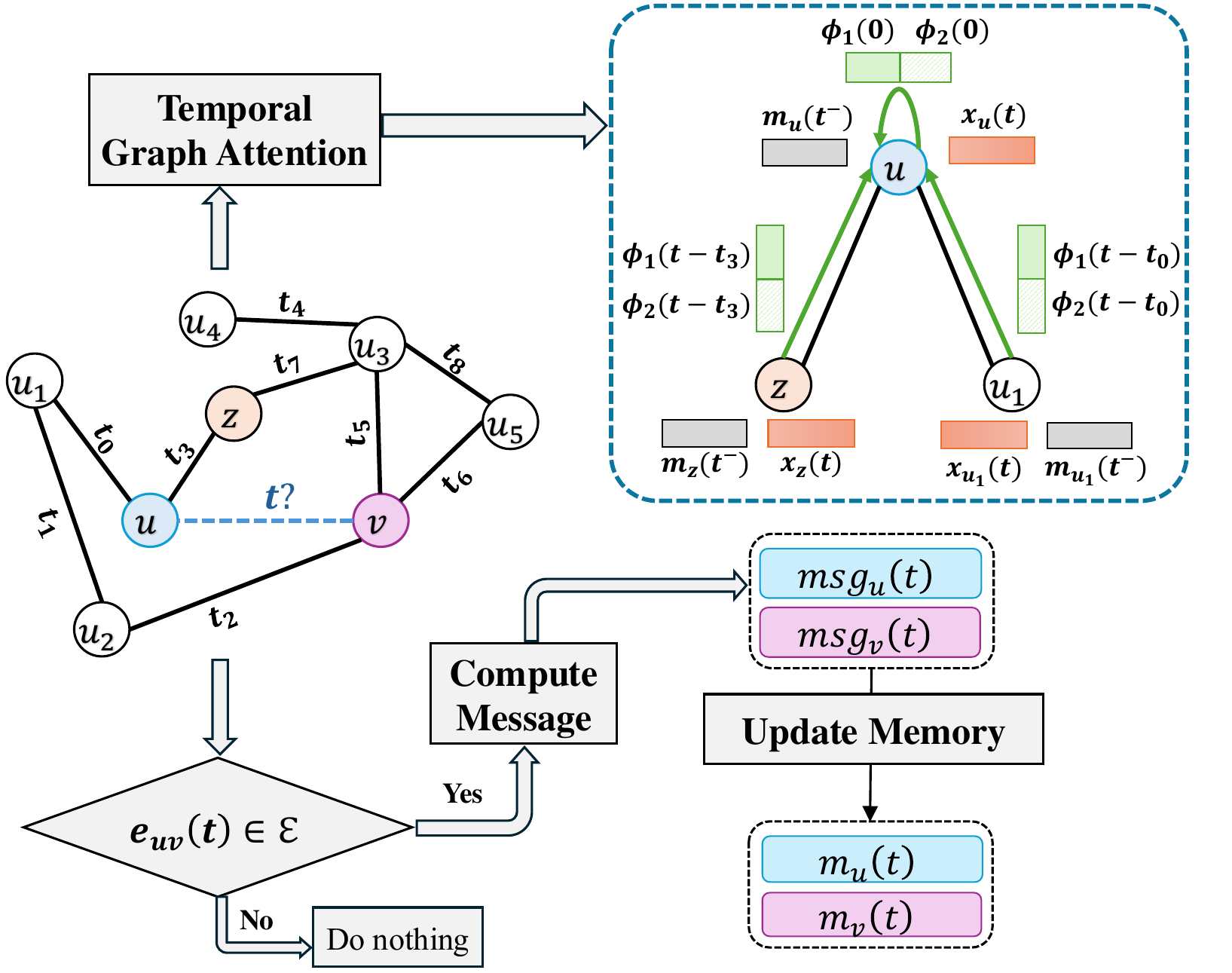}
    \vspace{-2pt}
    \caption{Overview of the MAE component, demonstrating its operation on an example graph.}
    \label{fig:mae_overview}

\end{figure}

To further improve the quality of latent representations, we incorporate the memory of nodes into the embedding module. A series of \( L \) graph attention layers is employed to compute node embeddings. Figure~\ref{fig:mae_overview} provides an overview of the Memory-Augmented Embedding (MAE) component, illustrating how memory updates are performed and how memory vectors enhance embeddings within the temporal graph attention mechanism. To reduce computational overhead, only the \( k \) most recent neighbors of each node are sampled for convolution. Specifically, for a node \( u \) with a set of neighbors \( N_u(t) = \{u_1, \dots, u_N\} \) and corresponding edge timestamps \( t_1, \dots, t_N \), we sample the \( k \) most recent neighbors as \( \{u_{N-k+1}, \dots,u_N\} \). The computations for layers \( l = 0, \dots, L-1 \) are as follows:
\begin{align*}
    h_u^{(0)}(t) &= m_u(t^-) + x_u(t), \\
    K^{(l)}(t) &= \begin{bmatrix}
        h_{u_j}^{(l-1)}(t) \Vert e_{uu_j}(t_j) \Vert \phi_1(t-t_j) \Vert \phi_2(t-t_j)
    \end{bmatrix}_{j=N-k+1}^N, \\
    V^{(l)}(t) &= K^{(l)}(t),\\
    q^{(l)}(t) &= h_u^{(l-1)}(t) \Vert \phi_1(0) \Vert \phi_2(0),\\
    \tilde{h}_u^{(l)}(t) &= 
    \text{MultiHeadAttention}^{(l)}\left( q^{(l)}(t), K^{(l)}(t), V^{(l)}(t) \right),\\
    h_u^{(l)}(t) &= \text{MLP}^{(l)}\left( h_u^{(l-1)}(t) \Vert \tilde{h}_u^{(l)}(t) \right).
\end{align*}

Here, \( \phi_1 \) is the time encoder defined earlier, while \( \phi_2 \) is a fixed, non-learnable time encoder shown to improve training stability~\cite{cong2023we}. The term \( h_{u_j}^{(l-1)}(t) \) refers to the latent representation of neighboring node \( u_j \) at layer \( l-1 \) and time \( t \). The initial embedding \( h_u^{(0)}(t) \) is computed by combining the node's memory vector \( m_u(t) \) with its feature vector \( x_u(t) \). MLP is a multi-layer perceptron known as a fully connected neural network as well. Subsequent layers refine the embeddings using attention mechanisms and temporal encoders, leveraging both structural and temporal information in the graph. The final embedding of node \( u \) is obtained from the last layer as:
\begin{equation}
h_u(t) = h_u^{(L-1)}(t). \label{eq:memory}
\end{equation}

\subsection{Neighbor Co-occurrence Embedding (NCE)}
\begin{figure}[htb]
\vspace{-1mm}
    \centering
    \includegraphics[width=0.89\columnwidth]{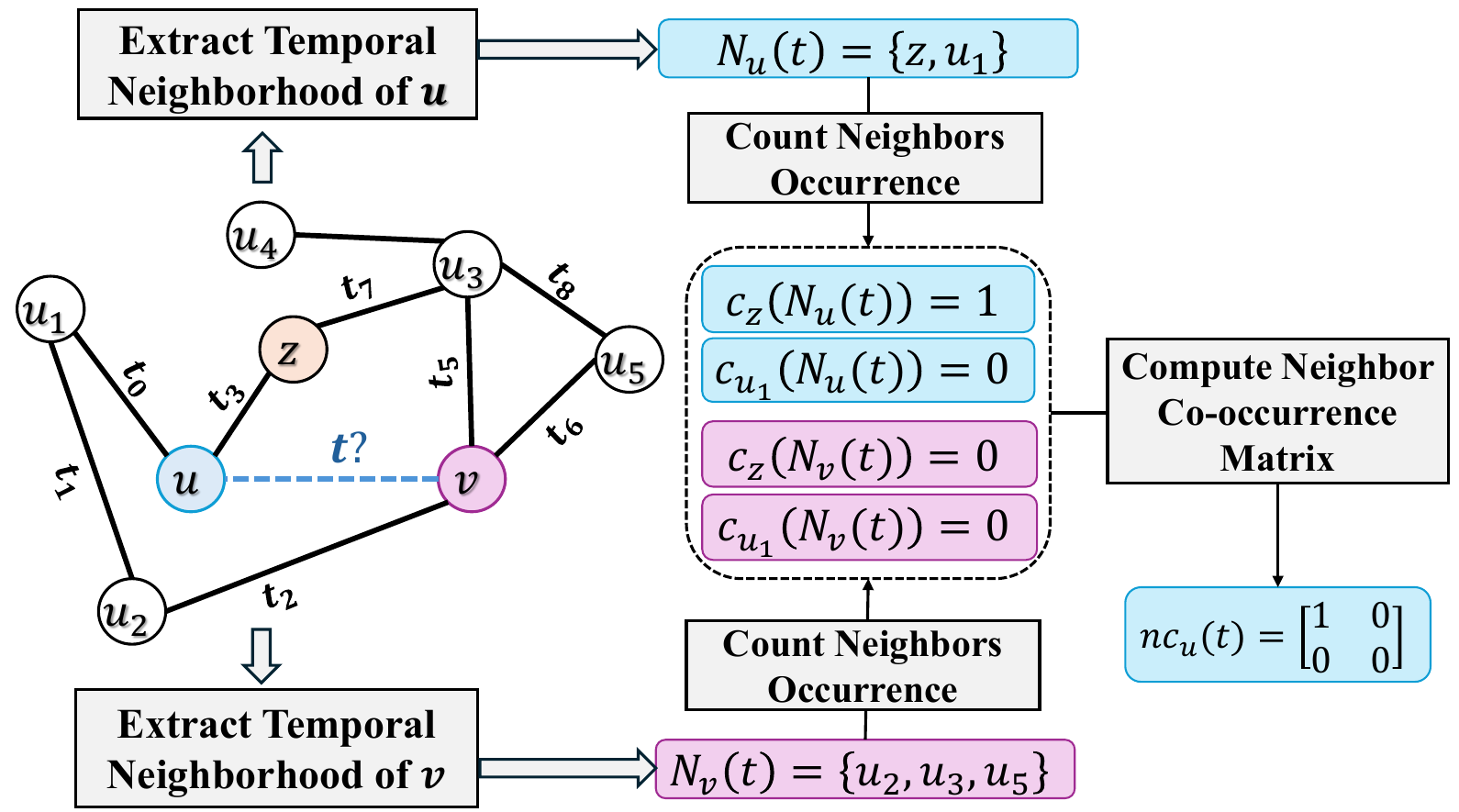}
    \vspace{-4pt}
    \caption{Illustration of \( nc_u(t) \) computation for node \( u \) in an example graph.}
    \label{fig:nce_example}
    \vspace{-2mm}
\end{figure}

Yu et al.~\cite{yu2023towards} demonstrated that considering the co-occurrence of nodes in the first-hop neighborhood significantly enhances the quality of embeddings. Inspired by their method, we consider the following approach to encode the first-hop neighborhood around a pair of nodes. We define the neighbor co-occurrence matrix as:
\begin{equation}
    \text{nc}_u(t) = \left[ c_z(N_u(t)), c_z(N_v(t)) \right]_{z \in N_u(t)},\nonumber
\end{equation}
where \( c_z(N_u(t)) \) is the number of times \( z \) appears in \( N_u(t) \), and \( c_z(N_v(t)) \) is the number of times \( z \) appears in \( N_v(t) \). Consequently, \( \text{nc}^t_u \) is a matrix with dimensions \( |N_u(t)| \times 2 \). Since the temporal neighborhood of nodes can be very large, we limit our focus to the \( r \) most recent neighbors of a node, where \( r \) is a hyperparameter. If a node has fewer than \( r \) neighbors, we pad the rows of the matrix with zeros to ensure a consistent size of \( r \times 2 \). Figure~\ref{fig:nce_example} illustrates this process, showing how temporal neighborhoods are extracted, neighbor occurrences are counted, and the matrix is constructed.

Next, we compute the co-occurrence embedding using the following equation:
\begin{equation}
    ce_u(t) = \text{MLP}(\text{nc}_u(t)[:, 0]) + \text{MLP}(\text{nc}_u(t)[:, 1]), \label{eq:cooc}
\end{equation}
where \( \text{nc}_u(t)[:, 0] \) and \( \text{nc}_u(t)[:, 1] \) represent the first and second columns of \( \text{nc}_u(t) \), respectively. The output of the MLP is a vector of size \( d_{ce} \), resulting in a matrix of size \( r \times d_{ce} \) before summation.

Finally, to obtain a consistent embedding representation, we flatten \( ce_u(t) \) into a column vector with dimensions \( r \cdot d_{ce} \times 1 \), preserving all positional information.

\subsection{Temporal Anonymous Walks with Restart (TAWR)}
\label{subsec:tawr}
Wang et al.~\cite{wang2021inductive} introduced \textit{causal anonymous walks} for inductive representation learning, where they performed multiple temporal walks from nodes \( u \) and \( v \) to capture causality in the network. In each temporal walk, links are sampled with a probability proportional to \( \exp(\alpha(t - t_p)) \), where \( \alpha \) is a non-negative hyperparameter, \( t \) is the timestamp of the current link, and \( t_p \) is the timestamp of the previously sampled link. Larger values of \( \alpha \) place greater emphasis on recent links, while \( \alpha = 0 \) results in uniform sampling across links. To ensure inductive learning, node identities are then replaced with hitting counts derived from the sampled walks, allowing the model to establish correlations between temporal motifs without relying on specific node labels.

Building on this approach, we propose \textbf{Temporal Anonymous Walks with Restart (TAWR)} to model interaction dynamics in temporal graphs. Interaction networks, such as user-product networks, often exhibit two types of user behavior: frequent revisitation of previously interacted items (\textit{exploitation}) and interaction with new items (\textit{exploration}). Traditional temporal walks do not explicitly distinguish between these behaviors, leading to suboptimal representations in dynamic settings. To address this, we introduce a probabilistic restart mechanism that enables the walk to revisit its starting node before continuing, ensuring a balance between exploration and exploitation.

\begin{definition}
    A \textit{temporal walk with restart} (TWR) from node \( u \) to node \( v \) in a temporal network is a temporal walk that may restart back to the originating node \( u \) based on a restart probability \( pr_u \). Given a random value \( p \in [0, 1] \), the TWR is defined as:
\[
{WR}_{u,v} =
\begin{cases} 
    {W}_{u,v}, & \text{if } p \leq pr_u, \\
    [{W}_{u,u_{k+1}}, (u_{k+1}, u, t_0), {W}_{u,v}], & \text{if } p > pr_u.
\end{cases}
\]

Here, \( W_{u,u_{k+1}} \) denotes the initial segment of the walk, \( (u_{k+1}, u, t_0) \) represents a virtual edge modeling the restart back to \( u \), and \( W_{u,v} \) corresponds to the remaining segment of the walk. Importantly, \( (u_{k+1}, u, t_0) \) does not have to be an actual edge in the graph but serves as a mechanism to simulate the restart process.
\end{definition}

This mechanism ensures that nodes exhibiting exploitation tendencies remain in the vicinity of their previous interactions, reinforcing local structures, while also allowing exploration beyond their immediate neighborhood when needed. By balancing these two aspects, TWRs provide a more nuanced approach to modeling dynamic interactions.

To capture the tendency of a node toward exploitation or revisitation, we make the restart probability \( pr_u(t) \) a learnable scalar parameter. This probability, representing the likelihood of a node restarting its walk, is trained in an end-to-end fashion along with the rest of the model. We initialize this scalar by applying an MLP on the embedding of node $u$ generated in Equation~\ref{eq:memory}:
\begin{equation}
pr_u(t) = \text{MLP}(h_u(t)). \label{eq:restart}
\end{equation}

\begin{figure}[htb]
    \centering
    \includegraphics[width=0.85\columnwidth]{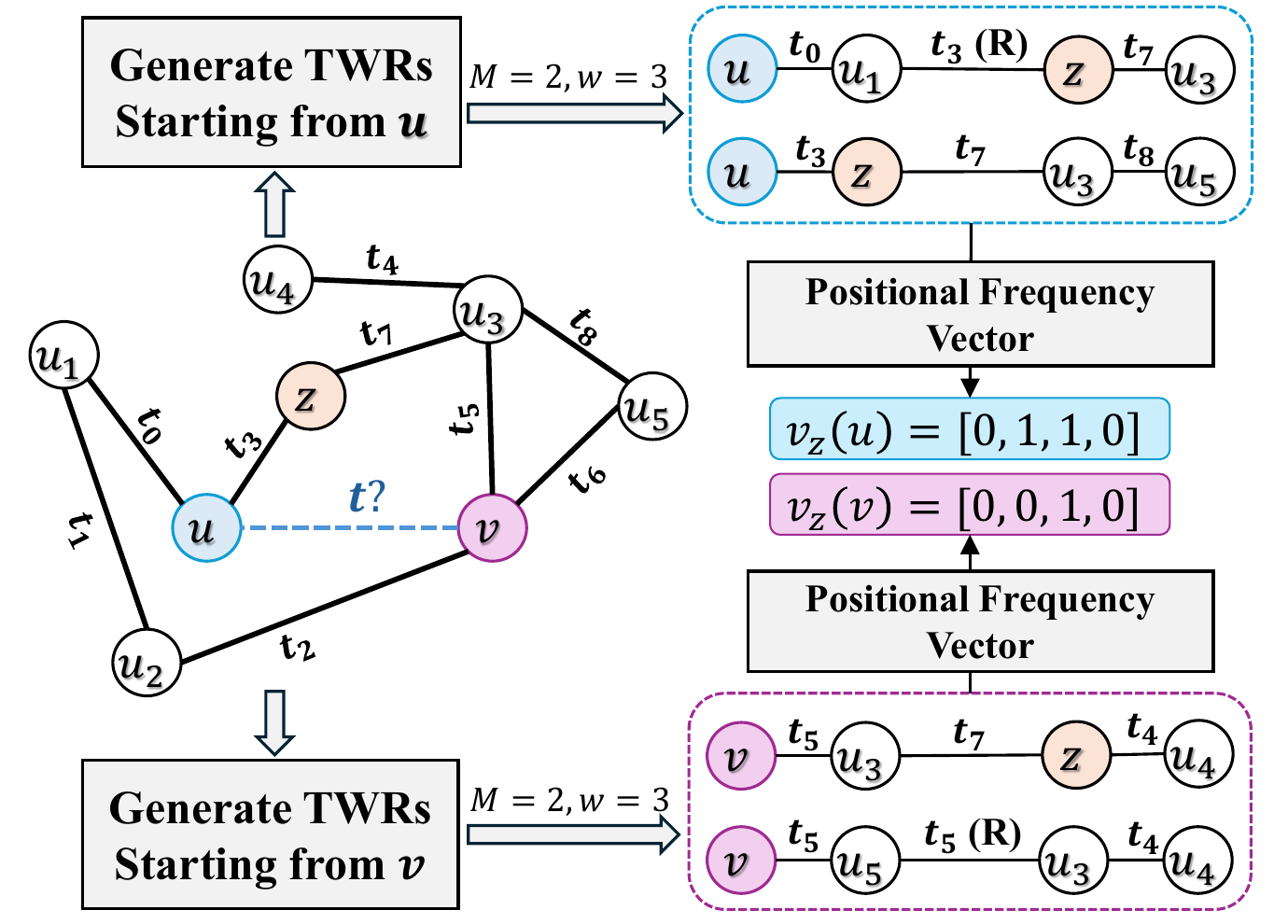}
   
    \caption{Illustration of the positional frequency vector computation for node \( z \) in an example graph.}
    \label{fig:tawr_walks}
    
\end{figure}
Next, we perform \( M \) temporal walks with restart from each node \( u \) and \( v \), with each walk having a length \( w \). In total, this results in \( 2M \) walks (\( WR_0, WR_1, \dots, WR_{2M} \)). To anonymize these walks, we adopt the set-based anonymization method from~\cite{wang2021inductive}. For a node \( z \), the \textit{positional frequency vector} \( \mathbf{v}_z(u) \) is defined based on the walks originating from node \( u \) as:
\[
\mathbf{v}_z(u) = [v_z^1, v_z^2, \dots, v_z^{w+1}],
\]
where \( v_z^p \) is the frequency of node \( z \) appearing at position \( p \) across all \( M \) walks starting from node \( u \). Figure~\ref{fig:tawr_walks} provides an illustration of this process, showcasing how positional frequency vectors are constructed based on sampled TWRs from nodes $u$ and $v$ in an example graph. The relative identity of \( z \) under our TAWR framework is represented as:
\[
I_{TAWR}(z) = \{\mathbf{v}_z(u), \mathbf{v}_z(v)\}.
\]

Using this representation, each node in a TWR is replaced with its relative identity \( I_{TAWR}(z) \). To encode the positional frequency vectors, we apply an MLP to each vector \( \mathbf{v}_z \). The resulting embeddings are then summed:
\[
E_z = \text{MLP}(\mathbf{v}_z(u)) + \text{MLP}(\mathbf{v}_z(v)),
\]
where \( E_z \) represents the encoded identity of node \( z \). The anonymized TWR is then expressed as:
\[\hat{WR_i} = [(E_{u_0}, E_{u_1}, t_0), (E_{u_1}, E_{u_2}, t_1), \dots, (E_{u_w}, E_{u_{w+1}}, t_{w+1})].\]

Each walk is encoded using a sequence encoder, such as an RNN, and for node $u$, the encoded walks are aggregated as follows:
\begin{align}
    enc(\hat{WR_i}) =&\, \text{RNN}([E_{u_0} \oplus \phi_1(0), E_{u_1} \oplus \phi_1(t_0 - t_1), \nonumber \\ 
    &\dots, E_{u_{w+1}} \oplus \phi_1(t_{w} - t_{w+1})]),\nonumber\\
    q_i &= enc(\hat{WR_i})^T, \nonumber\\
    K = V &= \left[ enc(\hat{WR_1}), enc(\hat{WR_2}), \dots, enc(\hat{WR_{M}}) \right], \nonumber\\
    enc(\hat{{WR}_{u}}) &= \text{MultiHeadAttention}(q_i, K, V). \label{eq:walk}
\end{align}
where $\phi_1$ is the time encoder defined in Section~\ref{subsec:mae}, and $enc(\hat{{WR}_{u}}) \in \mathbb{R}^{d_w}$.

\subsection{Classifier}
Each of the our proposed method's components generates embeddings for nodes in the temporal graph. To construct the final embedding for node \( u \), we concatenate the embeddings obtained from these components:
\begin{equation}
  emb_{u}(t) = [h_u(t) \Vert ce_u(t) \Vert enc(\hat{WR_{u}})\Vert pr_u(t)],  
\end{equation}
where \( h_u(t) \), \( ce_u(t) \), $enc(\hat{WR_{u}})$, and $pr_u(t)$ represent the embeddings produced in Equations~\eqref{eq:memory},~\eqref{eq:cooc},~\eqref{eq:restart}, and~\eqref{eq:walk}, respectively.

For the downstream task of link prediction, we concatenate the embeddings of the two nodes \( u \) and \( v \), and pass them through an MLP to compute the likelihood of a link forming between them at time \( t \):
\begin{equation}
    \hat{y}_{uv}(t) = \text{MLP}([\text{emb}_u(t) \Vert \text{emb}_v(t)]),
\end{equation}
where \( \hat{y}_{uv}(t) \) represents the predicted probability of a link existing between nodes \( u \) and \( v \) at time \( t \).

For node classification, we predict the label of node \( u \) at time \( t \) by passing its embedding through an MLP:
\begin{equation}
    \hat{y}_{u}(t) = \text{MLP}(\text{emb}_u(t)),
\end{equation}
where \( \hat{y}_{u}(t) \) is a probability distribution over \( c \) possible classes, representing the likelihood of node \( u \) belonging to each class at time \( t \). 

\vspace{-4pt}
\section{Experiments}
\label{sec:experiments}

In this section, we evaluate the effectiveness of \textbf{TAWRMAC} through extensive experiments. First, we outline the experimental settings (Section~\ref{sec:expsettings}). Next, we present the main results, comparing TAWRMAC with state-of-the-art methods on dynamic link prediction and node classification (Section~\ref{sec:results}). We then report computational efficiency (Section~\ref{subsec:computational_efficiency}), analyze the learnable restart in TAWR (Section~\ref{subsec:learnable_restart}), and study hyperparameter sensitivity (Section~\ref{subsec:hyperparam_sensitivity}). Finally, we conduct an ablation study to assess the contributions of individual components (Section~\ref{sec:ablation}).

\subsection{Experimental Settings}
\label{sec:expsettings}
In our experiments, we follow the DyGLib~\cite{yu2023towards} framework for consistent evaluation of dynamic graph learning models. Below, we outline the datasets, baselines, evaluation tasks, metrics, and implementation details.
\subsubsection{Datasets}
We evaluate our model on twelve diverse datasets:

\begin{table*}[htb]
\centering
\caption{Statistics of the Datasets. (E-Feat: Dimension of Edge Features)}
\label{tab:dataset_stats}
\begin{tabular}{l|ccccccc}
\hline
\textbf{Dataset} & \textbf{Domain}&  \textbf{\#Nodes} & \textbf{\#Edges} & \textbf{E-Feat}  & \textbf{\#Time Steps} & \textbf{Time Granularity} & \textbf{Time Span} \\
\hline
Wikipedia       & Web & 9,227  & 157,474     & 172   & 152,757 & Unix timestamps & 1 month \\
Reddit          & Social Media & 10,984 & 672,447     & 172   & 669,065 & Unix timestamps & 1 month \\
MOOC            & Education & 7,144  & 411,749     & 4     & 345,600 & Unix timestamps & 17 months \\
LastFM          & Music &1,980  & 1,293,103   & --    & 1,283,614 & Unix timestamps & 1 month \\
Enron           & Commerce & 184    & 125,235     & --    & 22,632  & Unix timestamps & 3 years \\
UCI             &Social Network & 1,899  & 59,835      & --    & 58,911  & Unix timestamps & 196 days \\
Flights         & Transportation &13,169 & 1,927,145   & 1     & 122     & Days & 4 months \\
Can. Parl.      & Politics & 734    & 74,478      & 1     & 14      & Years & 14 years \\
US Legis.       & Politics &225    & 60,396      & 1     & 12      & Congresses & 12 congresses \\
UN Trade        & Commerce & 255    & 507,497     & 1     & 32      & Years & 32 years \\
UN Vote         & Politics & 201    & 1,035,742   & 1     & 72      & Years & 72 years \\
Contact         &  Proximity Network& 692    & 2,426,279   & 1     & 8,064   & 5 minutes & 1 month \\
\hline
\end{tabular}
\end{table*}

\begin{itemize}
    \item \textbf{Wikipedia}: This dataset captures interactions between users and Wikipedia pages over a one-month period. Nodes represent users and pages, while edges denote editing actions with associated timestamps. Each edge is annotated with a 172-dimensional feature vector derived from the Linguistic Inquiry and Word Count (LIWC) tool~\cite{pennebaker2001linguistic}. This dataset also includes node labels that determine whether a user is banned.

    \item \textbf{Reddit}: This dataset tracks user activity on subreddits for a one-month duration. Nodes correspond to users and subreddits, while edges represent timestamped posting events. Similar to Wikipedia, each edge is accompanied by a 172-dimensional LIWC feature vector, and nodes labels determine whether a user is banned from the platform.

    \item \textbf{MOOC}: A bipartite network that documents interactions between students and online course content, such as problem sets and videos. Edges indicate a student's access to specific content, with each edge containing a 4-dimensional feature vector.

    \item \textbf{LastFM}: A music interaction dataset where nodes represent users and songs, and edges capture the timestamped \emph{user-listens-to-so ng} relationships over a one-month period. The dataset contains no additional edge features.

    \item \textbf{Enron}: A dataset of email exchanges among employees of the Enron corporation spanning three years. Nodes represent employees, and edges correspond to timestamped emails. No additional attributes are provided for nodes or edges.


    \item \textbf{UCI}: An online communication network among University of California students. Nodes represent individuals, and edges correspond to timestamped messages exchanged between them.

    \item \textbf{Flights}: A dynamic network capturing air traffic patterns during the COVID-19 pandemic. Nodes represent airports, and edges denote flights between them. Each edge is weighted by the number of flights per day.

    \item \textbf{Canadian Parliament (Can. Parl.)}: A political interaction network recording voting behaviors of Canadian Members of Parliament (MPs) from 2006 to 2019. Nodes represent MPs, and edges are formed when two MPs vote ``yes'' on the same bill, with edge weights indicating the frequency of such votes in a year.

    \item \textbf{US Legislature (US Legis.)}: A senate co-sponsorship network documenting collaboration among U.S. senators. Nodes represent senators, and edges are weighted by the number of times two senators co-sponsored a bill during a congressional session.

    \item \textbf{UN Trade}: A temporal trade network focusing on food and agriculture exchanges among 181 nations over 30 years. Nodes represent countries, and edges are weighted by the total normalized trade volume (import or export) between countries.

    \item \textbf{UN Vote}: A dataset tracking roll-call voting patterns in the United Nations General Assembly from 1946 to 2020. Nodes represent nations, and edges are weighted by the number of times two nations both voted ``yes'' on an item.

    \item \textbf{Contact}: This dataset captures the temporal evolution of physical proximity among 700 university students over four weeks. Nodes are students, and edges represent close-proximity events, with edge weights indicating the strength of physical proximity.
\end{itemize}


These datasets were collected by Poursafaei et al.~\cite{poursafaei2022towards}, and are publicly available under MIT license or Apache Lincense 2.0.\footnote{\url{https://zenodo.org/record/7213796\#.Y1cO6y8r30o}}
Table~\ref{tab:dataset_stats} shows the statistics of these datasets.

\subsubsection{Baselines}
We evaluate our approach against ten established baselines: JODIE~\cite{kumar2019predicting}, DyRep~\cite{trivedi2019dyrep}, TGAT~\cite{xu2020inductive}, TGN~\cite{rossi2020temporal}, CAWN~\cite{wang2021inductive}, EdgeBank~\cite{poursafaei2022towards}, TCL~\cite{wang2021tcl}, GraphMixer~\cite{cong2023we}, DyGFormer~\cite{yu2023towards}, and FreeDyG~\cite{tian2024freedyg}. To ensure consistency and minimize computational overhead, we use the reported results from the DyGFormer, and FreeDyG papers maintaining alignment in library versions and random seeds and ensuring the same test sets. In Section~\ref{sec:related_work}, we reviewed these methods.

\subsubsection{Evaluation}

We consider two key tasks for evaluating the models:

\noindent \textbf{ (I) Dynamic Link Prediction:} This task involves predicting the likelihood of a link forming between two nodes at a specific time. We evaluate models under two important settings:

\begin{itemize}
    \item In the \textit{transductive setting}, the model predicts future links between nodes seen during training, leveraging prior interactions and embeddings. For example, given their known histories, it might predict whether two social network users will connect. 
    \item The \textit{inductive setting} involves predicting links for unseen nodes, making it more challenging as the model must generalize to new nodes. This scenario is crucial for dynamic networks with evolving structures, such as predicting interactions for new users or items in a recommender system.

\end{itemize}

These settings ensure comprehensive evaluation, testing both the model's refinement of known embeddings and its adaptability to unseen data.
The evaluation uses Average Precision (AP)~\cite{zhu2004recall} and Area Under the Receiver Operating Characteristic Curve (AUC-ROC) as metric. Three negative sampling strategies (NSS), including random, historical, and inductive~\cite{poursafaei2022towards}, are applied to ensure a comprehensive evaluation. In random NSS, negative edges are selected randomly while ensuring no overlap with positive edges. In historical NSS, negative edges are drawn from positive edges that occurred in previous timestamps but are missing in the current step. In inductive NSS, negative edges are chosen from positive edges that appeared during test time in earlier timestamps but are absent at the current step. For further details, please refer to~\cite{poursafaei2022towards}.

\noindent \textbf{(II) Dynamic Node Classification:} This task predicts the label or state of a node during an interaction. Due to label imbalance in the datasets, we use AUC-ROC as the evaluation metric.

For both tasks, the datasets are chronologically split into 70\%/15\%/15\% for training, validation, and testing, ensuring temporal consistency in the evaluation process.

\subsubsection{Implementation Details} We optimized TAWRMAC using Adam optimization function~\cite{diederik2014adam} with the learning rate set to $0.0001$, and a batch size of 200. We trained TAWRMAC for 100 epochs with early stopping based on validation loss with patience of 5. We ran the method five times with seeds from 0 to 4 and report the average performance to eliminate deviations. The experiments were conducted on a system with an AMD EPYC 7513 CPU, 4 NVIDIA RTX A4000 GPUs, and 1 TB of RAM.
Table~\ref{tab:exp_settings} lists the configuration we keep fixed across all datasets, and Table~\ref{tab:extra_hyper_params} reports the dataset-specific settings tuned on validation. When multiple values appear in a cell, they correspond to random/historical/inductive negative sampling, respectively (see table footnote).
\begin{table}[H]
\caption{Configuration of TAWRMAC across all datasets.}
\label{tab:exp_settings}
\setlength{\tabcolsep}{2pt}
\scalebox{0.9}{
\begin{tabular}{c|l|c}
\hline
\textbf{Module} & \textbf{Hyper-parameter} & \textbf{Value} \\\hline
\multirow{2}{*}{All} &
Dropout & 0.1\\
&Learnable time encoding dimension ($d_{\phi_1}$)&  100 \\
\hline
\multirow{5}{*}{MAE} &
Fixed time encoding dimension ($d_{\phi_2}$) &  20\\
&Memory dimension ($d_m$) &  172 \\
&\# sampled neighbors for graph convolution ($k$) & 10 \\
&\# of graph attention heads &   2\\
&\# of graph convolution layers ($L$) & 1\\
\hline
\multirow{1}{*}{NCE} &
Neighbor co-occurrence encoding dimension ($d_{ce}$) &  10\\
\hline
\multirow{3}{*}{TAWR} &
\# of attention heads for walk encoding &   4\\
&Length of each walk ($w$) &    4\\
&Time scaling factor ($\alpha$) & $1\times10^{-6}$\\
\hline
\end{tabular}}
\end{table}

\begin{table}[H]
\centering
\caption{Configuration of number of temporal walks with restart ($\bm{M}$), number of sampled neighbors in NCE ($\bm{r}$), positional frequency vector dimension ($\bm{d_v}$), walk encoding dimension ($\bm{d_w}$), and neighbor sampling strategies (\textit{NbSS}) across different datasets.}
\setlength{\tabcolsep}{3pt}
\label{tab:extra_hyper_params}
\begin{tabular}{l|ccccc}
\hline
\textbf{Dataset} & $\bm{M}$ & $\bm{r}$ & $\bm{d_v}$ & $\bm{d_w}$ &\textbf{NbSS}  \\
\hline
Wikipedia*       & 10/10/1  & 32/32/128 & 100/100/10 & 172/172/10 & recent    \\
Reddit          & 10 & 32 & 100 &172 &  recent      \\
MOOC            & 10  & 32  & 100 &172 &  recent    \\
LastFM*          & 10  & 32/32/4 &100 &  172 & recent    \\
Enron*           & 10/1/1  & 32/8/8 & 100/10/10 & 172/10/10 &  recent      \\
UCI*             & 10  & 32/4/4 & 100 &172 &  recent      \\
Flights         & 20 & 32 &100 &  172 & recent    \\
Can. Parl. & 30    & 500 &100 &  172 & uniform     \\
US Legis.*  & 10    & 200/32/32 & 100 &172 &  recent       \\
UN Trade        & 20    & 200 &100 & 172 &  uniform      \\
UN Vote         & 20    & 100 &100 &  172 & uniform  \\
Contact         & 10    & 32 &100 &  172 & recent   \\
\hline
\end{tabular}
\begin{flushleft}
\footnotesize{* For columns with multiple values, the first value corresponds to random NS, the second to historical NS, and the third to inductive NS.}
\end{flushleft}
\end{table}

\subsection{Experimental Results}
\label{sec:results}

\begin{table*}[htb]
\centering

\caption{Experimental results for inductive dynamic link prediction using the Average Precision (AP) metric.}

\label{tab:ind_ap_comparison}
\setlength{\tabcolsep}{2pt}
\scalebox{0.9}{
\begin{tabular}{p{14pt}|c|c|c|c|c|c|c|c|c|c}
\hline
\textbf{NSS}       & \textbf{Dataset} & \textbf{JODIE} & \textbf{DyRep} & \textbf{TGAT} & \textbf{TGN} & \textbf{CAWN} & \textbf{TCL} & \textbf{GraphMixer} & \textbf{DyGFormer} & \textbf{TAWRMAC} \\ 
\hline
\multirow{13}{*}{\rotatebox{-90}{Random}} 
& Wikipedia  & 94.82 $\pm$ 0.20 & 92.43 $\pm$ 0.37 & 96.22 $\pm$ 0.07 & 97.83 $\pm$ 0.04 & 98.24 $\pm$ 0.03 & 96.22 $\pm$ 0.17 & 96.65 $\pm$ 0.02 & \underline{98.59 $\pm$ 0.03} & \textbf{98.93 $\pm$ 0.03} \\
& Reddit     & 96.50 $\pm$ 0.13 & 96.09 $\pm$ 0.11 & 97.09 $\pm$ 0.04 & 97.50 $\pm$ 0.07 & 98.62 $\pm$ 0.01 & 94.09 $\pm$ 0.07 & 95.26 $\pm$ 0.02 & \underline{98.84 $\pm$ 0.02} & \textbf{98.99 $\pm$ 0.05} \\
& MOOC       & 79.63 $\pm$ 1.92 & 81.07 $\pm$ 0.44 & 85.50 $\pm$ 0.19 & \underline{89.04 $\pm$ 1.17} & 81.42 $\pm$ 0.24 & 80.60 $\pm$ 0.22 & 81.41 $\pm$ 0.21 & 86.96 $\pm$ 0.43 & \textbf{91.14 $\pm$ 0.82} \\
& LastFM     & 81.61 $\pm$ 3.82 & 83.02 $\pm$ 1.48 & 78.63 $\pm$ 0.31 & 81.45 $\pm$ 4.29 & 89.42 $\pm$ 0.07 & 73.53 $\pm$ 1.66 & 82.11 $\pm$ 0.42 & \textbf{94.23 $\pm$ 0.09} & \underline{93.07 $\pm$ 1.37} \\
& Enron      & 80.72 $\pm$ 1.39 & 74.55 $\pm$ 3.95 & 67.05 $\pm$ 1.51 & 77.94 $\pm$ 1.02 & 86.35 $\pm$ 0.51 & 76.14 $\pm$ 0.79 & 75.88 $\pm$ 0.48 & \textbf{89.76 $\pm$ 0.34} & \underline{89.45 $\pm$ 0.12} \\
& UCI        & 79.86 $\pm$ 1.48 & 57.48 $\pm$ 1.87 & 79.54 $\pm$ 0.48 & 88.12 $\pm$ 2.05 & 92.73 $\pm$ 0.06 & 87.36 $\pm$ 2.03 & 91.19 $\pm$ 0.42 & \underline{94.54 $\pm$ 0.12} & \textbf{95.08 $\pm$ 0.33} \\
& Flights    & 94.74 $\pm$ 0.37 & 92.88 $\pm$ 0.73 & 88.73 $\pm$ 0.33 & 95.03 $\pm$ 0.60 & 97.06 $\pm$ 0.02 & 83.41 $\pm$ 0.07 & 83.03 $\pm$ 0.05 & \textbf{97.79 $\pm$ 0.02} & \underline{97.37 $\pm$ 0.12} \\
& Can. Parl. & 53.92 $\pm$ 0.94 & 54.02 $\pm$ 0.76 & 55.18 $\pm$ 0.79 & 54.10 $\pm$ 0.93 & 55.80 $\pm$ 0.69 & 54.30 $\pm$ 0.66 & \underline{55.91 $\pm$ 0.82} & \textbf{87.74 $\pm$ 0.71} & 55.90 $\pm$ 0.83 \\
& US Legis.  & 54.93 $\pm$ 2.29 & 57.28 $\pm$ 0.71 & 51.00 $\pm$ 3.11 & \underline{58.63 $\pm$ 0.37} & 53.17 $\pm$ 1.20 & 52.59 $\pm$ 0.97 & 50.71 $\pm$ 0.76 & 54.28 $\pm$ 2.87 & \textbf{59.28 $\pm$ 1.12} \\
& UN Trade   & 59.65 $\pm$ 0.77 & 57.02 $\pm$ 0.69 & 61.03 $\pm$ 0.18 & 58.31 $\pm$ 3.15 & \textbf{65.24 $\pm$ 0.21} & 62.21 $\pm$ 0.12 & 62.17 $\pm$ 0.31 & \underline{64.55 $\pm$ 0.62} & 63.86 $\pm$ 1.83 \\
& UN Vote    & 56.64 $\pm$ 0.96 & 54.62 $\pm$ 2.22 & 52.24 $\pm$ 1.46 & \textbf{58.85 $\pm$ 2.51} & 49.94 $\pm$ 0.45 & 51.60 $\pm$ 0.97 & 50.68 $\pm$ 0.44 & 55.93 $\pm$ 0.39 & \underline{58.16 $\pm$ 1.63} \\
& Contact    & 94.34 $\pm$ 1.45 & 92.18 $\pm$ 0.41 & 95.87 $\pm$ 0.11 & 93.82 $\pm$ 0.99 & 89.55 $\pm$ 0.30 & 91.11 $\pm$ 0.12 & 90.59 $\pm$ 0.05 & \underline{98.03 $\pm$ 0.02} & \textbf{98.23 $\pm$ 0.06} \\\cline{2-11}
& Avg. Rank  & 6 & 6.75 & 6.25 & 4.5 & 4.33 & 6.92 & 6.33 & \underline{2.17} & \textbf{1.67} \\
\Hline

\multirow{13}{*}{\rotatebox{-90}{Historical}} 
& Wikipedia  & 68.69 $\pm$ 0.39 & 62.18 $\pm$ 1.27 & 84.17 $\pm$ 0.22 & 81.76 $\pm$ 0.32 & 67.27 $\pm$ 1.63 & 82.20 $\pm$ 2.18 & \textbf{87.60 $\pm$ 0.30} & 71.42 $\pm$ 4.43 & \underline{84.37 $\pm$ 0.25} \\
& Reddit     & 62.34 $\pm$ 0.54 & 61.60 $\pm$ 0.72 & 63.47 $\pm$ 0.36 & 64.85 $\pm$ 0.85 & 63.67 $\pm$ 0.41 & 60.83 $\pm$ 0.25 & 64.50 $\pm$ 0.26 & \underline{65.37 $\pm$ 0.68} & \textbf{67.95 $\pm$ 0.65} \\
& MOOC       & 63.22 $\pm$ 1.55 & 62.93 $\pm$ 1.24 & 76.73 $\pm$ 0.29 & 77.07 $\pm$ 3.41 & 74.68 $\pm$ 0.68 & 74.27 $\pm$ 0.53 & 74.00 $\pm$ 0.97 & \textbf{80.82 $\pm$ 0.30} & \underline{79.78 $\pm$ 4.67} \\
& LastFM     & 70.39 $\pm$ 4.31 & 71.45 $\pm$ 1.76 & 76.27 $\pm$ 0.25 & 66.65 $\pm$ 6.11 & 71.33 $\pm$ 0.47 & 65.78 $\pm$ 0.65 & \underline{76.42 $\pm$ 0.22} & 76.35 $\pm$ 0.52 & \textbf{76.53 $\pm$ 0.23} \\
& Enron      & 65.86 $\pm$ 3.71 & 62.08 $\pm$ 2.27 & 61.40 $\pm$ 1.31 & 62.91 $\pm$ 1.16 & 60.70 $\pm$ 0.36 & 67.11 $\pm$ 0.62 & \underline{72.37 $\pm$ 1.37} & 67.07 $\pm$ 0.62 & \textbf{77.25 $\pm$ 0.77} \\
& UCI        & 63.11 $\pm$ 2.27 & 52.47 $\pm$ 2.06 & 70.52 $\pm$ 0.93 & 70.78 $\pm$ 0.78 & 64.54 $\pm$ 0.47 & 76.71 $\pm$ 1.00 & \underline{81.66 $\pm$ 0.49} & 72.13 $\pm$ 1.87 & \textbf{83.48 $\pm$ 0.15} \\
& Flights    & 61.01 $\pm$ 1.65 & 62.83 $\pm$ 1.31 & \underline{64.72 $\pm$ 0.36} & 59.31 $\pm$ 1.43 & 56.82 $\pm$ 0.57 & 64.50 $\pm$ 0.25 & \textbf{65.28 $\pm$ 0.24} & 57.11 $\pm$ 0.21 & 55.77 $\pm$ 0.33 \\
& Can. Parl. & 52.60 $\pm$ 0.88 & 52.28 $\pm$ 0.31 & 56.72 $\pm$ 0.47 & 54.42 $\pm$ 0.77 & 57.14 $\pm$ 0.07 & 55.71 $\pm$ 0.74 & 55.84 $\pm$ 0.73 & \textbf{87.40 $\pm$ 0.85} & \underline{57.76 $\pm$ 0.33} \\
& US Legis.  & 52.94 $\pm$ 2.11 & \textbf{62.10 $\pm$ 1.41} & 51.83 $\pm$ 3.95 & \underline{61.18 $\pm$ 1.10} & 55.56 $\pm$ 1.71 & 53.87 $\pm$ 1.41 & 52.03 $\pm$ 1.02 & 56.31 $\pm$ 3.46 & 58.41 $\pm$ 2.18 \\
& UN Trade   & 55.46 $\pm$ 1.19 & 55.49 $\pm$ 0.84 & 55.28 $\pm$ 0.71 & 52.80 $\pm$ 3.19 & 55.00 $\pm$ 0.38 & \underline{55.76 $\pm$ 1.03} & 54.94 $\pm$ 0.97 & 53.20 $\pm$ 1.07 & \textbf{56.51 $\pm$ 3.13} \\
& UN Vote    & 61.04 $\pm$ 1.30 & 60.22 $\pm$ 1.78 & 53.05 $\pm$ 3.10 & \underline{63.74 $\pm$ 3.00} & 47.98 $\pm$ 0.84 & 54.19 $\pm$ 2.17 & 48.09 $\pm$ 0.43 & 52.63 $\pm$ 1.26 & \textbf{65.54 $\pm$ 0.85} \\
& Contact    & 90.42 $\pm$ 2.34 & 89.22 $\pm$ 0.66 & \textbf{94.15 $\pm$ 0.45} & 88.13 $\pm$ 1.50 & 74.20 $\pm$ 0.80 & 90.44 $\pm$ 0.17 & 89.91 $\pm$ 0.36 & \underline{93.56 $\pm$ 0.52} & 93.32 $\pm$ 0.10 \\\cline{2-11}
& Avg. Rank  & 6.17  & 6.25  & 4.83  & 5.33  & 6.67  & 5.00  & 4.42  & \underline{4.08} & \textbf{2.25} \\

\Hline
\multirow{13}{*}{\rotatebox{-90}{Inductive}} 
& Wikipedia  & 68.70 $\pm$ 0.39 & 62.19 $\pm$ 1.28 & 84.17 $\pm$ 0.22 & 81.77 $\pm$ 0.32 & 67.24 $\pm$ 1.63 & 82.20 $\pm$ 2.18 & \textbf{87.60 $\pm$ 0.29} & 71.42 $\pm$ 4.43 & \underline{86.12 $\pm$ 1.24} \\
& Reddit     & 62.32 $\pm$ 0.54 & 61.58 $\pm$ 0.72 & 63.40 $\pm$ 0.36 & 64.84 $\pm$ 0.84 & 63.65 $\pm$ 0.41 & 60.81 $\pm$ 0.26 & 64.49 $\pm$ 0.25 & \underline{65.35 $\pm$ 0.82} & \textbf{65.49 $\pm$ 0.77} \\
& MOOC       & 63.22 $\pm$ 1.55 & 62.92 $\pm$ 1.24 & 76.72 $\pm$ 0.30 & 77.07 $\pm$ 3.40 & 74.69 $\pm$ 0.68 & 74.28 $\pm$ 0.53 & 73.99 $\pm$ 0.97 & \underline{80.82 $\pm$ 0.30} & \textbf{81.43 $\pm$ 2.85} \\
& LastFM     & 70.39 $\pm$ 4.31 & 71.45 $\pm$ 1.75 & 76.28 $\pm$ 0.25 & 69.46 $\pm$ 4.65 & 71.33 $\pm$ 0.47 & 65.78 $\pm$ 0.65 & \underline{76.42 $\pm$ 0.22} & 76.35 $\pm$ 0.52 & \textbf{77.47 $\pm$ 0.26} \\
& Enron      & 65.86 $\pm$ 3.71 & 62.08 $\pm$ 2.27 & 61.40 $\pm$ 1.30 & 62.90 $\pm$ 1.16 & 60.72 $\pm$ 0.36 & 67.11 $\pm$ 0.62 & \underline{72.37 $\pm$ 1.38} & 67.07 $\pm$ 0.62 & \textbf{77.28 $\pm$ 0.59} \\

& UCI & 63.16 $\pm$ 2.27 & 52.47 $\pm$ 2.09 & 70.49 $\pm$ 0.93 & 70.73 $\pm$ 0.79 & 64.54 $\pm$ 0.47 & 76.65 $\pm$ 0.99 & \underline{81.64 $\pm$ 0.49} & 72.13 $\pm$ 1.86 & \textbf{82.85 $\pm$ 0.65} \\

& Flights & 61.01 $\pm$ 1.66 & 62.83 $\pm$ 1.31 & \underline{64.72 $\pm$ 0.37} & 59.32 $\pm$ 1.45 & 56.82 $\pm$ 0.56 & 64.50 $\pm$ 0.25 & \textbf{65.29 $\pm$ 0.24} & 57.11 $\pm$ 0.20 & 55.38 $\pm$ 0.82 \\
& Can. Parl. & 52.58 $\pm$ 0.86 & 52.24 $\pm$ 0.28 & 56.46 $\pm$ 0.50 & 54.18 $\pm$ 0.73 & 57.06 $\pm$ 0.08 & 55.46 $\pm$ 0.69 & 55.76 $\pm$ 0.65 & \textbf{87.22 $\pm$ 0.82} & \underline{58.32 $\pm$ 0.90} \\ 

& US Legis. & 52.94 $\pm$ 2.11 & \textbf{62.10 $\pm$ 1.41} & 51.83 $\pm$ 3.95 & \underline{61.18 $\pm$ 1.10} & 55.56 $\pm$ 1.71 & 53.87 $\pm$ 1.41 & 52.03 $\pm$ 1.02 & 56.31 $\pm$ 3.46 & 58.87 $\pm$ 2.77 \\
& UN Trade & 55.43 $\pm$ 1.20 & 55.42 $\pm$ 0.87 & 55.58 $\pm$ 0.68 & 52.80 $\pm$ 3.24 & 54.97 $\pm$ 0.38 & \underline{55.66 $\pm$ 0.98} & 54.88 $\pm$ 1.01 & 52.56 $\pm$ 1.70 & \textbf{55.98 $\pm$ 2.15} \\
& UN Vote & 61.17 $\pm$ 1.33 & 60.29 $\pm$ 1.79 & 53.08 $\pm$ 3.10 & \underline{63.71 $\pm$ 2.97} & 48.01 $\pm$ 0.82 & 54.13 $\pm$ 2.16 & 48.10 $\pm$ 0.40 & 52.61 $\pm$ 1.25 & \textbf{66.71 $\pm$ 2.15} \\
& Contact & 90.43 $\pm$ 2.33 & 89.22 $\pm$ 0.65 & \textbf{94.14 $\pm$ 0.45} & 88.12 $\pm$ 1.50 & 74.19 $\pm$ 0.81 & 90.43 $\pm$ 0.17 & 89.91 $\pm$ 0.36 & 93.55 $\pm$ 0.52 & \underline{93.71 $\pm$ 0.45} \\ \cline{2-11}
& Avg. Rank & 6.08 & 6.42 & 4.67 & 5.25 & 6.67 & 5.00 & 4.42 & \underline{4.33} & \textbf{2.08} \\\hline
\end{tabular}
}
\end{table*}
Table~\ref{tab:ind_ap_comparison} presents the performance of various methods on the AP metric for inductive dynamic link prediction across three negative sampling strategies. It is important to note that EdgeBank is exclusively designed for transductive dynamic link prediction and, therefore, its performance under the inductive setting is not included. For clarity and improved readability, the reported results are scaled by a factor of 100. From the table, it is evident that TAWRMAC excels in experiments using the random negative sampling strategy, achieving first place in 6 out of 12 datasets and consistently ranking among the top three models with an impressive average rank of 1.67. In the historical and inductive negative sampling experiments, TAWRMAC demonstrates robust performance across most datasets, except for the Flights dataset, where it faces some challenges. Nevertheless, it remains among the top three methods in all other datasets, achieving the best average ranks of 2.25 and 2.08 for the historical and inductive settings, respectively. This highlights the overall effectiveness and adaptability of TAWRMAC across various evaluation scenarios. Similarly, Table~\ref{tab:trans_ap_comparison} reports AP results for transductive dynamic link prediction, where TAWRMAC again outperforms all baselines, achieving the best average ranks of 1.58, 1.67, and 2.5 for random, historical, and inductive sampling, respectively. Results using the AUC-ROC metric showed similar trends, and are presented in Section~\ref{sec:exp_results_appendix} in the Appendix.
\begin{table*}[htb]
\vspace{-2mm}
\centering
\caption{Experimental results for transductive dynamic link prediction using the Average Precision (AP) metric.}
\label{tab:trans_ap_comparison}
\setlength{\tabcolsep}{2pt}

\scalebox{0.9}{
\begin{tabular}{p{14pt}|c|c|c|c|c|c|c|c|c|c|c}
\hline
\textbf{NSS}       & \textbf{Dataset} & \textbf{JODIE} & \textbf{DyRep} & \textbf{TGAT} & \textbf{TGN} & \textbf{CAWN} & \textbf{EdgeBank} & \textbf{TCL} & \textbf{GraphMixer} & \textbf{DyGFormer} & \textbf{TAWRMAC} \\ 
\hline
\multirow{13}{*}{\rotatebox{-90}{Random}} 
& Wikipedia  & 96.5 $\pm$ 0.14  & 94.86 $\pm$ 0.06 & 96.94 $\pm$ 0.06 & 98.45 $\pm$ 0.06 & 98.76 $\pm$ 0.03 & 90.37 $\pm$ 0.00 & 96.47 $\pm$ 0.16 & 97.25 $\pm$ 0.03 & \underline{99.03 $\pm$ 0.02} & \textbf{99.29 $\pm$ 0.02} \\
& Reddit     & 98.31 $\pm$ 0.14 & 98.22 $\pm$ 0.04 & 98.52 $\pm$ 0.02 & 98.63 $\pm$ 0.06 & 99.11 $\pm$ 0.01 & 94.86 $\pm$ 0.00 & 97.53 $\pm$ 0.02 & 97.31 $\pm$ 0.01 & \underline{99.22 $\pm$ 0.01} & \textbf{99.3 $\pm$ 0.01}  \\
& MOOC       & 80.23 $\pm$ 2.44 & 81.97 $\pm$ 0.49 & 85.84 $\pm$ 0.15 & \underline{89.15 $\pm$ 1.60} & 80.15 $\pm$ 0.25 & 57.97 $\pm$ 0.00 & 82.38 $\pm$ 0.24 & 82.78 $\pm$ 0.15 & 87.52 $\pm$ 0.49 & \textbf{91.46 $\pm$ 0.66} \\
& LastFM     & 70.85 $\pm$ 2.13 & 71.92 $\pm$ 2.21 & 73.42 $\pm$ 0.21 & 77.07 $\pm$ 3.97 & 86.99 $\pm$ 0.06 & 79.29 $\pm$ 0.00 & 67.27 $\pm$ 2.16 & 75.61 $\pm$ 0.24 & \textbf{93.00 $\pm$ 0.12} & \underline{91.17 $\pm$ 0.08} \\
& Enron      & 84.77 $\pm$ 0.30 & 82.38 $\pm$ 3.36 & 71.12 $\pm$ 0.97 & 86.53 $\pm$ 1.11 & 89.56 $\pm$ 0.09 & 83.53 $\pm$ 0.00 & 79.70 $\pm$ 0.71 & 82.25 $\pm$ 0.16 & \underline{92.47 $\pm$ 0.12} & \textbf{92.74 $\pm$ 0.21} \\
& UCI        & 89.43 $\pm$ 1.09 & 65.14 $\pm$ 2.30 & 79.63 $\pm$ 0.70 & 92.34 $\pm$ 1.04 & 95.18 $\pm$ 0.06 & 76.20 $\pm$ 0.00 & 89.57 $\pm$ 1.63 & 93.25 $\pm$ 0.57 & \underline{95.79 $\pm$ 0.17} & \textbf{96.40 $\pm$ 0.15}  \\
& Flights    & 95.60 $\pm$ 1.73 & 95.29 $\pm$ 0.72 & 94.03 $\pm$ 0.18 & 97.95 $\pm$ 0.14 & 98.51 $\pm$ 0.01 & 89.35 $\pm$ 0.00 & 91.23 $\pm$ 0.02 & 90.99 $\pm$ 0.05 & \textbf{98.91 $\pm$ 0.01} & \underline{98.84 $\pm$ 0.93} \\
& Can. Parl. & 69.26 $\pm$ 0.31 & 66.54 $\pm$ 2.76 & 70.73 $\pm$ 0.72 & 70.88 $\pm$ 2.34 & 69.82 $\pm$ 2.34 & 64.55 $\pm$ 0.00 & 68.67 $\pm$ 2.67 & \underline{77.04 $\pm$ 0.46} & \textbf{97.36 $\pm$ 0.45} & 76.42 $\pm$ 1.47 \\
& US Legis.  & 75.05 $\pm$ 1.52 & \underline{75.34 $\pm$ 0.39} & 68.52 $\pm$ 3.16 & \textbf{75.99 $\pm$ 0.58} & 70.58 $\pm$ 0.48 & 58.39 $\pm$ 0.00 & 69.59 $\pm$ 0.48 & 70.74 $\pm$ 1.02 & 71.11 $\pm$ 0.59 & 73.89 $\pm$ 0.52 \\
& UN Trade   & 64.94 $\pm$ 0.31 & 63.21 $\pm$ 0.93 & 61.47 $\pm$ 0.18 & 65.03 $\pm$ 1.37 & 65.39 $\pm$ 0.12 & 60.41 $\pm$ 0.00 & 62.21 $\pm$ 0.03 & 62.61 $\pm$ 0.27 & \underline{66.46 $\pm$ 1.29} & \textbf{68.11 $\pm$ 0.9} \\
& UN Vote    & 63.91 $\pm$ 0.81 & 62.81 $\pm$ 0.80 & 52.21 $\pm$ 0.98 & \underline{65.72 $\pm$ 2.17} & 52.84 $\pm$ 0.10 & 58.49 $\pm$ 0.00 & 51.90 $\pm$ 0.30 & 52.11 $\pm$ 0.16 & 55.55 $\pm$ 0.42 & \textbf{66.5 $\pm$ 1.17} \\
& Contact    & 95.31 $\pm$ 1.33 & 95.98 $\pm$ 0.15 & 96.28 $\pm$ 0.09 & 96.89 $\pm$ 0.56 & 90.26 $\pm$ 0.28 & 92.58 $\pm$ 0.00 & 92.44 $\pm$ 0.12 & 91.92 $\pm$ 0.03 & \underline{98.29 $\pm$ 0.01} & \textbf{98.80 $\pm$ 0.03}  \\\cline{2-12}
& Avg. Rank  & 5.92  & 6.67  & 6.83  & 3.5   & 5     & 8.42  & 8.08  & 6.58  & \underline{2.42}  & \textbf{1.58}  \\

\Hline
\multirow{13}{*}{\rotatebox{-90}{Historical}} 
& Wikipedia  & 83.01 $\pm$ 0.66 & 79.93 $\pm$ 0.56 & 87.38 $\pm$ 0.22 & 86.86 $\pm$ 0.33 & 71.21 $\pm$ 1.67 & 73.35 $\pm$ 0.00 & 89.05 $\pm$ 0.39 & \textbf{90.9 $\pm$ 0.10}  & 82.23 $\pm$ 2.54 & \underline{90.16 $\pm$ 0.27} \\
& Reddit     & 80.03 $\pm$ 0.36 & 79.83 $\pm$ 0.31 & 79.55 $\pm$ 0.20 & 81.22 $\pm$ 0.61 & 80.82 $\pm$ 0.45 & 73.59 $\pm$ 0.00 & 77.14 $\pm$ 0.16 & 78.44 $\pm$ 0.18 & \underline{81.57 $\pm$ 0.67} & \textbf{84.36 $\pm$ 0.43} \\
& MOOC       & 78.94 $\pm$ 1.25 & 75.6 $\pm$ 1.12 & 82.19 $\pm$ 0.62 & \underline{87.06 $\pm$ 1.93} & 74.05 $\pm$ 0.95 & 60.71 $\pm$ 0.00 & 77.06 $\pm$ 0.41 & 77.77 $\pm$ 0.92 & 85.85 $\pm$ 0.66 & \textbf{88.93 $\pm$ 0.39} \\
& LastFM     & 74.35 $\pm$ 3.81 & 74.92 $\pm$ 2.46 & 71.59 $\pm$ 0.24 & 76.87 $\pm$ 4.64 & 69.86 $\pm$ 0.43 & 73.03 $\pm$ 0.00 & 59.3 $\pm$ 2.31 & 72.47 $\pm$ 0.49 & \underline{81.57 $\pm$ 0.48} & \textbf{82.69 $\pm$ 0.56} \\
& Enron      & 69.85 $\pm$ 2.70 & 71.19 $\pm$ 2.76 & 64.07 $\pm$ 1.05 & 73.91 $\pm$ 1.76 & 64.73 $\pm$ 0.36 & 76.53 $\pm$ 0.00 & 70.66 $\pm$ 0.39 & \underline{77.98 $\pm$ 0.92} & 75.63 $\pm$ 0.73 & \textbf{80.23 $\pm$ 0.79} \\
& UCI        & 75.24 $\pm$ 5.80 & 55.1 $\pm$ 3.14 & 68.27 $\pm$ 1.37 & 80.43 $\pm$ 2.12 & 65.3 $\pm$ 0.43 & 65.5 $\pm$ 0.00 & 80.25 $\pm$ 2.74 & \underline{84.11 $\pm$ 1.35} & 82.17 $\pm$ 0.82 & \textbf{88.29 $\pm$ 0.14} \\
& Flights    & 66.48 $\pm$ 2.59 & 67.61 $\pm$ 0.99 & \textbf{72.38 $\pm$ 0.18} & 66.7 $\pm$ 1.64 & 64.72 $\pm$ 0.97 & 70.53 $\pm$ 0.00 & 70.68 $\pm$ 0.24 & \underline{71.47 $\pm$ 0.26} & 66.59 $\pm$ 0.49 & 68.57 $\pm$ 0.56 \\
& Can. Parl. & 51.79 $\pm$ 0.63 & 63.31 $\pm$ 1.23 & 67.13 $\pm$ 0.84 & 68.42 $\pm$ 3.07 & 66.53 $\pm$ 2.77 & 63.84 $\pm$ 0.00 & 65.93 $\pm$ 3.00 & 74.34 $\pm$ 0.87 & \textbf{97 $\pm$ 0.31}    & \underline{76.6 $\pm$ 2.38}  \\
& US Legis.  & 51.71 $\pm$ 5.76 & \underline{86.88 $\pm$ 2.25} & 62.14 $\pm$ 6.60 & 74 $\pm$ 7.57 & 68.82 $\pm$ 8.23 & 63.22 $\pm$ 0.00 & 80.53 $\pm$ 3.95 & 81.65 $\pm$ 1.02 & 85.3 $\pm$ 3.88 & \textbf{91.17 $\pm$ 2.58} \\
& UN Trade   & 61.39 $\pm$ 1.83 & 59.19 $\pm$ 1.07 & 55.74 $\pm$ 0.91 & 58.44 $\pm$ 5.51 & 55.71 $\pm$ 0.38 & \textbf{81.32 $\pm$ 0.00} & 55.9 $\pm$ 1.17 & 57.05 $\pm$ 1.22 & 64.41 $\pm$ 1.40 & \underline{67.99 $\pm$ 2.68} \\
& UN Vote    & 70.02 $\pm$ 0.81 & 69.3 $\pm$ 1.12 & 52.96 $\pm$ 2.14 & 69.37 $\pm$ 3.93 & 51.26 $\pm$ 0.04 & \textbf{84.89 $\pm$ 0.00} & 52.3 $\pm$ 2.35 & 51.2 $\pm$ 1.60 & 60.84 $\pm$ 1.58 & \underline{74.67 $\pm$ 2.58} \\
& Contact    & 95.31 $\pm$ 2.13 & 96.39 $\pm$ 0.20 & 96.05 $\pm$ 0.52 & 93.05 $\pm$ 2.35 & 84.16 $\pm$ 0.49 & 88.81 $\pm$ 0.00 & 93.86 $\pm$ 0.21 & 93.36 $\pm$ 0.41 & \underline{97.57 $\pm$ 0.06} & \textbf{97.81 $\pm$ 0.09} \\
\cline{2-12}
& Avg. Rank  & 6.33  & 6     & 6.25  & 4.75  & 8.5   & 6.42  & 6.5   & 4.92  & \underline{3.67}  & \textbf{1.67}  \\

\Hline
\multirow{13}{*}{\rotatebox{-90}{Inductive}}
& Wikipedia  & 75.65 $\pm$ 0.79 & 70.21 $\pm$ 1.58 & 87 $\pm$ 0.16 & 85.62 $\pm$ 0.44 & 74.06 $\pm$ 2.62 & 80.63 $\pm$ 0.00 & 86.76 $\pm$ 0.72 & \underline{88.59 $\pm$ 0.17} & 78.29 $\pm$ 5.38 & \textbf{89.55 $\pm$ 1.37} \\
& Reddit     & 86.98 $\pm$ 0.16 & 86.3 $\pm$ 0.26 & 89.59 $\pm$ 0.24 & 88.1 $\pm$ 0.24 & \textbf{91.67 $\pm$ 0.24} & 85.48 $\pm$ 0.00 & 87.45 $\pm$ 0.29 & 85.26 $\pm$ 0.11 & \underline{91.11 $\pm$ 0.40} & 90.8 $\pm$ 0.28  \\
& MOOC       & 65.23 $\pm$ 2.19 & 61.66 $\pm$ 0.95 & 75.95 $\pm$ 0.64 & 77.5 $\pm$ 2.91 & 73.51 $\pm$ 0.94 & 49.43 $\pm$ 0.00 & 74.65 $\pm$ 0.54 & 74.27 $\pm$ 0.92 & \underline{81.24 $\pm$ 0.69} & \textbf{83.42 $\pm$ 2.57} \\
& LastFM     & 62.67 $\pm$ 4.49 & 64.41 $\pm$ 2.70 & 71.13 $\pm$ 0.17 & 65.95 $\pm$ 5.98 & 67.48 $\pm$ 0.77 & \textbf{75.49 $\pm$ 0.00} & 58.21 $\pm$ 0.89 & 68.12 $\pm$ 0.33 & \underline{73.97 $\pm$ 0.50} & 72.57 $\pm$ 0.37 \\
& Enron      & 68.96 $\pm$ 0.98 & 67.79 $\pm$ 1.53 & 63.94 $\pm$ 1.36 & 70.89 $\pm$ 2.72 & 75.15 $\pm$ 0.58 & 73.89 $\pm$ 0.00 & 71.29 $\pm$ 0.32 & 75.01 $\pm$ 0.79 & \underline{77.41 $\pm$ 0.89} & \textbf{78.58 $\pm$ 0.59} \\
& UCI        & 65.99 $\pm$ 1.40 & 54.79 $\pm$ 1.76 & 68.67 $\pm$ 0.84 & 70.94 $\pm$ 0.71 & 64.61 $\pm$ 0.48 & 57.43 $\pm$ 0.00 & 76.01 $\pm$ 1.11 & \underline{80.1 $\pm$ 0.51}  & 72.25 $\pm$ 1.71 & \textbf{81.46 $\pm$ 0.9} \\

& Flights    & 69.07 $\pm$ 4.02 & 70.57 $\pm$ 1.82 & \underline{75.48 $\pm$ 0.26} & 71.09 $\pm$ 2.72 & 69.18 $\pm$ 1.52 & \textbf{81.08 $\pm$ 0.00} & 74.62 $\pm$ 0.18 & 74.87 $\pm$ 0.21 & 70.92 $\pm$ 1.78 & 65.62 $\pm$ 1.25 \\
& Can. Parl. & 48.42 $\pm$ 0.66 & 58.61 $\pm$ 0.86 & 68.82 $\pm$ 1.21 & 65.34 $\pm$ 2.87 & 67.75 $\pm$ 1.00 & 62.16 $\pm$ 0.00 & 65.85 $\pm$ 1.75 & 69.48 $\pm$ 0.63 & \textbf{95.44 $\pm$ 0.57} & \underline{69.87 $\pm$ 1.8} \\
& US Legis.  & 50.27 $\pm$ 5.13 & \underline{83.44 $\pm$ 1.16} & 61.91 $\pm$ 5.82 & 67.57 $\pm$ 6.47 & 65.81 $\pm$ 8.52 & 64.74 $\pm$ 0.00 & 78.15 $\pm$ 3.34 & 79.63 $\pm$ 0.84 & 81.25 $\pm$ 3.62 & \textbf{84.02 $\pm$ 1.98} \\
& UN Trade   & 60.42 $\pm$ 1.48 & 60.19 $\pm$ 1.24 & 60.61 $\pm$ 1.24 & 61.04 $\pm$ 6.01 & \underline{62.54 $\pm$ 0.67} & \textbf{72.97 $\pm$ 0.00} & 61.06 $\pm$ 1.74 & 60.15 $\pm$ 1.29 & 55.79 $\pm$ 1.02 & 60.64 $\pm$ 1.26 \\
& UN Vote    & \underline{67.79 $\pm$ 1.46} & 67.53 $\pm$ 1.98 & 52.89 $\pm$ 1.61 & 67.63 $\pm$ 2.67 & 52.19 $\pm$ 0.34 & 66.3 $\pm$ 0.00 & 50.62 $\pm$ 0.82 & 51.6 $\pm$ 0.73 & 51.91 $\pm$ 0.84 & \textbf{69.63 $\pm$ 1.98} \\
& Contact    & 93.43 $\pm$ 1.78 & 94.18 $\pm$ 0.10 & 94.35 $\pm$ 0.48 & 90.18 $\pm$ 3.28 & 89.31 $\pm$ 0.27 & 85.2 $\pm$ 0.00 & 91.35 $\pm$ 0.21 & 90.87 $\pm$ 0.35 & \underline{94.75 $\pm$ 0.28} & \textbf{96.01 $\pm$ 0.1} \\
\cline{2-12}
& Avg. Rank  & 7.5   & 7.33  & 5.08     & 5.42  & 6     & 6.08  & 5.67  & 5.33  & \underline{4.08}  & \textbf{2.5}  \\ 

\hline
\end{tabular}
}

\end{table*}

Table~\ref{tab:freedyg_results} presents a comparison between FreeDyG~\cite{tian2024freedyg} and TAWRMAC on dynamic link prediction tasks under three negative sampling strategies in both transductive and inductive settings. Since FreeDyG only reports results on 6 of our 12 datasets, we restrict our comparison to this subset and present it in a separate table to ensure a fair evaluation, avoiding skewed average rank comparisons due to differing dataset coverage. Across all settings, TAWRMAC consistently achieves equal or better average ranks than FreeDyG. In particular, it demonstrates strong performance under challenging settings like historical and inductive negative sampling, where higher-order and stable temporal modeling is critical—areas where FreeDyG's first-hop-only encoding may fall short. 

\begin{table}[H]
\centering
\caption{AP results for transductive and inductive dynamic link prediction, comparing FreeDyG and TAWRMAC.}
\label{tab:freedyg_results}
\setlength{\tabcolsep}{2pt}
\renewcommand{\arraystretch}{1.2}
\scalebox{0.9}{
\begin{tabular}{p{14pt}|l|c|c||c|c}
\hline
\textbf{NSS} & \textbf{Dataset} & \multicolumn{2}{c||}{\textbf{Transductive}} & \multicolumn{2}{c}{\textbf{Inductive}} \\
\cline{3-6}
& & \textbf{FreeDyG} & \textbf{TAWRMAC} & \textbf{FreeDyG} & \textbf{TAWRMAC} \\
\hline
\multirow{7}{*}{\rotatebox{-90}{Random}}
  & Wikipedia & 99.26 ± 0.01 & \textbf{99.29 ± 0.02} & \textbf{98.97 ± 0.01} & 98.93 ± 0.03 \\
  & Reddit    & \textbf{99.48 ± 0.01} & 99.30 ± 0.01 & 98.91 ± 0.01 & \textbf{98.99 ± 0.01} \\
  & MOOC      & 89.61 ± 0.19 & \textbf{91.46 ± 0.66} & 87.75 ± 0.62 & \textbf{91.14 ± 0.82} \\
  & LastFM    & \textbf{92.15 ± 0.16} & 91.17 ± 0.08 & \textbf{94.89 ± 0.01} & 93.07 ± 0.05 \\
  & Enron     & 92.51 ± 0.05 & \textbf{92.74 ± 0.21} & \textbf{89.69 ± 0.17} & 89.45 ± 0.33 \\
  & UCI       & 96.28 ± 0.11 & \textbf{96.40 ± 0.15} & 94.85 ± 0.10 & \textbf{95.08 ± 0.12} \\
  \cline{2-6}
  & Avg. Rank & 1.67 & \textbf{1.33} & \textbf{1.50} & \textbf{1.50} \\
\hline
\multirow{7}{*}{\rotatebox{-90}{Historical}}
  & Wikipedia & \textbf{91.59 ± 0.57} & 90.16 ± 0.27 & 82.78 ± 0.30 & \textbf{84.37 ± 0.25} \\
  & Reddit    & \textbf{85.67 ± 1.01} & 84.36 ± 0.43 & 66.02 ± 0.41 & \textbf{67.95 ± 0.65} \\
  & MOOC      & 86.71 ± 0.81 & \textbf{88.93 ± 3.52} & \textbf{81.63 ± 0.33} & 79.78 ± 4.67 \\
  & LastFM    & 79.71 ± 0.51 & \textbf{82.69 ± 0.10} & \textbf{77.28 ± 0.21} & 76.53 ± 0.23 \\
  & Enron     & 78.87 ± 0.82 & \textbf{80.23 ± 0.79} & 73.01 ± 0.88 & \textbf{77.25 ± 0.77} \\
  & UCI       & 86.10 ± 1.19 & \textbf{88.29 ± 0.14} & 82.35 ± 0.39 & \textbf{83.48 ± 0.15} \\
  \cline{2-6}
  & Avg. Rank & 1.67 & \textbf{1.33} & 1.67 & \textbf{1.33} \\
\hline
\multirow{7}{*}{\rotatebox{-90}{Inductive}}
  & Wikipedia & \textbf{90.05 ± 0.79} & 89.55 ± 1.37 & \textbf{87.54 ± 0.26} & 86.12 ± 1.24 \\
  & Reddit    & 90.74 ± 0.17 & \textbf{90.80 ± 0.28} & 64.98 ± 0.20 & \textbf{65.49 ± 0.77} \\
  & MOOC      & 83.01 ± 0.87 & \textbf{83.42 ± 2.57} & 81.41 ± 0.31 & \textbf{81.43 ± 2.85} \\
  & LastFM    & 72.19 ± 0.24 & \textbf{72.57 ± 0.37} & 77.01 ± 0.43 & \textbf{77.47 ± 0.26} \\
  & Enron     & 77.81 ± 0.65 & \textbf{78.58 ± 0.59} & 72.85 ± 0.81 & \textbf{77.28 ± 0.59} \\
  & UCI       & \textbf{82.35 ± 0.73} & 81.46 ± 0.90 & 82.06 ± 0.58 & \textbf{82.85 ± 0.65} \\
  \cline{2-6}
  & Avg. Rank & 1.67 & \textbf{1.33} & 1.83 & \textbf{1.17} \\
\hline
\end{tabular}
}
\vspace{-2mm}
\end{table}

\vspace{-2mm}
\begin{table}[H]

\centering
\caption{AUC-ROC results for dynamic node classification.}
\label{tab:node_cl_results}
\begin{tabular}{l|cc|c}
\hline
\textbf{Method} & \textbf{Wikipedia} & \textbf{Reddit} & \textbf{Avg. Rank} \\
\hline
JODIE~\cite{kumar2019predicting}       & \textbf{88.99 $\pm$ 1.05} & 60.37 $\pm$ 2.58 & 4.50 \\
DyRep~\cite{trivedi2019dyrep}       & 86.39 $\pm$ 0.98 & 63.72 $\pm$ 1.32 & 5.00 \\
TGAT~\cite{xu2020inductive}        & 84.09 $\pm$ 1.27 & \underline{70.04 $\pm$ 1.09} & 4.00 \\
TGN~\cite{rossi2020temporal}         & 86.38 $\pm$ 2.34 & 63.27 $\pm$ 0.90 & 6.00 \\
CAWN~\cite{wang2021inductive}        & 84.88 $\pm$ 1.33 & 66.34 $\pm$ 1.78 & 5.00 \\
TCL~\cite{wang2021tcl}         & 77.83 $\pm$ 2.13 & 68.87 $\pm$ 2.15 & 5.00 \\
GraphMixer~\cite{cong2023we}  & 86.80 $\pm$ 0.79 & 64.22 $\pm$ 3.32 & 4.00 \\
DyGFormer~\cite{yu2023towards}   & 87.44 $\pm$ 1.08 & 68.00 $\pm$ 1.74 & \underline{2.50} \\
TAWRMAC        & \underline{87.69 $\pm$ 0.78} & \textbf{71.45 $\pm$ 0.92} & \textbf{1.50} \\
\hline
\end{tabular}
\end{table}
Table~\ref{tab:node_cl_results} reports the performance of baseline methods and our proposed model on dynamic node classification, evaluated on the Wikipedia and Reddit datasets—the only datasets that have dynamic node labels. FreeDyG is not included in this comparison, as it does not report results on node classification task. In this task, the predicted label for a user indicates whether they will be banned from editing (Wikipedia) or posting (Reddit). The results highlight the superior performance of our model, which achieves the best average rank of 1.5 among all compared methods. Notably, our model secures the top position on the Reddit dataset and attains a close second on the Wikipedia dataset, trailing the best-performing method by only 1.3 percentage points. 

\vspace{-4mm}
\subsection{Computational Efficiency}
\label{subsec:computational_efficiency}
To evaluate computational efficiency, we compare TAWRMAC against its ablation variant (w/o TAWR), CAWN (a walk-based model)~\cite{wang2021inductive}, and DyGFormer~\cite{yu2023towards} on three datasets: MOOC, UCI, and Wikipedia. For CAWN and DyGFormer, we used the official DyGLib implementation~\cite{yu2023towards} with optimal hyperparameters per dataset. We ran TAWRMAC and its ablated variant (w/o TAWR)  with the best configurations reported in our experiments (Section~\ref{sec:results}). The results, averaged over five runs, are shown in Table~\ref{tab:runtime_memory}.


\begin{table}[htb]
\centering
\caption{Runtime (min) and peak memory (MB) usage across five runs.}
\label{tab:runtime_memory}
\setlength{\tabcolsep}{2pt}
\renewcommand{\arraystretch}{1.2}
\scalebox{0.9}{
\begin{tabular}{l|l|c|c|c|c}
\hline
\textbf{Dataset} & \textbf{Measure} & \textbf{DyGFormer} & \textbf{CAWN} & \textbf{TAWRMAC} & \textbf{w/o TAWR} \\
\hline
\multirow{2}{*}{MOOC} 
& Runtime & 2,533.29 & 3,499.4 & 1,871.23 & 127.4 \\
& Memory  & 4,868    & 14,492   & 3,732    & 2,352   \\
\hline
\multirow{2}{*}{UCI} 
& Runtime & 712.50  & 419.58  & 53.01   & 7.27   \\
& Memory  & 2,536    & 14,948   & 3,210    & 1,660   \\
\hline
\multirow{2}{*}{Wikipedia} 
& Runtime & 1,000.53 & 522.05  & 44.47   & 30.96  \\
& Memory  & 2,684    & 10,050   & 2,616    & 2,510   \\
\hline
\end{tabular}
}
\vspace{-8mm}
\end{table}

TAWRMAC demonstrates significantly better efficiency than CAWN, particularly in memory usage. This improvement stems from CAWN’s reliance on an excessive number of temporal walks per node, which increases computational cost. In contrast, TAWRMAC achieves strong performance with far fewer walks, benefiting from its complementary modules (MAE and NCE). Compared to DyGFormer, TAWRMAC converges substantially faster during training, resulting in lower overall runtime—despite DyGFormer being marginally faster in inference.

\subsection{Analysis of Learnable Restart}
\label{subsec:learnable_restart}
In Section~\ref{subsec:tawr}, we introduced a node-specific, time-dependent restart probability $\textit{pr}(u, t)$, predicted via an MLP based on dynamically updated memory-augmented embeddings. This mechanism enables adaptive balancing between exploration and exploitation based on the temporal behavior of individual nodes. Now, we dig deeper into the learnable restart. 

\subsubsection{Behavioral Correlation Analysis}

To evaluate the interpretability of $\textit{pr}$, we analyze its correlation with node characteristics across 200 randomly selected test nodes in the UCI and Wikipedia datasets shown in Table~\ref{tab:spearman_correlation}. The results align with intuitive behavioral patterns:

\begin{itemize}
  \item \textbf{UCI}: High-degree users exhibit higher $\textit{pr}$ values, indicating a tendency toward exploration; infrequent users exhibit lower $\textit{pr}$ values, suggesting exploitative behavior.
  \item \textbf{Wikipedia}: High-degree editors exhibit lower $\textit{pr}$ values (exploitation), whereas infrequent editors tend to explore more, yielding higher $\textit{pr}$ values.
\end{itemize}

\begin{figure*}[htb]
    \centering
    \includegraphics[width=0.87\textwidth]{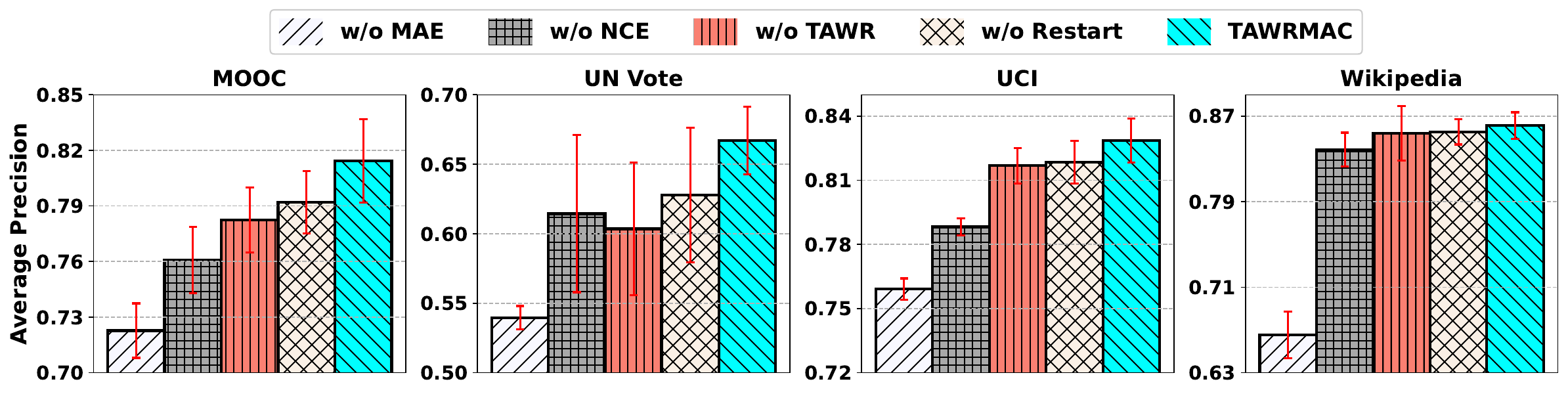}
    \vspace{-8pt}
    \caption{Ablation study results showcasing the performance (Average Precision) of TAWRMAC and its variants across four datasets (MOOC, UN Vote, UCI, and Wikipedia). Error bars indicate standard deviations.}
    \label{fig:ablation_study}
    \vspace{-2mm}
\end{figure*}

\begin{table}[htb]

\centering
\caption{Spearman correlation between $\textit{pr}$, node degree, and inter-event time.}
\setlength{\tabcolsep}{2pt}
\renewcommand{\arraystretch}{1.4}
\scalebox{0.9}{
\begin{tabular}{l|lccc}
\hline
\textbf{Dataset} & \textbf{Characteristic} & \textbf{Correlation} & \textbf{$p$-value} & \textbf{Significant} \\
\hline
\multirow{2}{*}{UCI} & Node Degree & 0.365 & $1.12 \times 10^{-7}$ & \checkmark \\
                     & Avg. Inter-event Time & -0.404 & $3.50 \times 10^{-9}$ & \checkmark \\
                     \hline
\multirow{2}{*}{Wikipedia} & Node Degree & -0.270 & $1.09 \times 10^{-4}$ & \checkmark \\
                           & Avg. Inter-event Time & 0.144 & $4.80 \times 10^{-2}$ & \checkmark \\
\hline
\end{tabular}
}
\label{tab:spearman_correlation}

\end{table}

\subsubsection{Comparison to Fixed Restart Strategies}
Here, we compare the learnable $\textit{pr}$ with fixed values and a degree-based heuristic defined as: 
\begin{align*}
    pr(u)=1-\frac{deg(u)}{max_v \deg(v)}
\end{align*}

\noindent This heuristic reflects the inverse relationship between node degree and restart probability, consistent with our empirical findings in the previous section. The results are included in Table~\ref{tab:pr_settings}. The learnable $\textit{pr}$ consistently outperforms fixed and heuristic alternatives, demonstrating its effectiveness in adapting to node-specific temporal dynamics. Additionally, based on our experiment on Wikipedia dataset, degree-based fixed $\textit{pr}$ achieves lower AP than the learnable variant, confirming the benefit of dynamic, time-dependent restart.
\vspace{-10pt}
\begin{table}[htb]
\centering
\caption{AP under different $\textit{pr}$ settings.}
\renewcommand{\arraystretch}{1.1}
\scalebox{0.9}{
\begin{tabular}{l|cccc}
\hline
\textbf{Dataset} & \textbf{Learnable \textit{pr}} & \textbf{$\textit{pr}=0.1$} & \textbf{$\textit{pr}=0.8$} & \textbf{Degree-based} \\
\hline
UCI & \textbf{82.85} & 81.81 & 81.53 & -- \\
Wikipedia & \textbf{86.12} & 81.93 & 85.00 & 82.55 $\pm$ 3.40 \\
\hline
\end{tabular}
}
\label{tab:pr_settings}
\end{table}
\vspace{-10pt}
\subsection{Hyperparameter Sensitivity}
\label{subsec:hyperparam_sensitivity}
We adopted standard hyperparameters from prior work and tuned critical ones, most importantly the number of temporal walks ($M$) and the number of sampled neighbors in the NCE module ($r$). A sensitivity analysis (demonstrated in Table~\ref{tab:sensitivity}) across three datasets shows that TAWRMAC achieves optimal performance with tuned values but remains relatively robust to suboptimal settings. All values were selected based on validation performance to balance accuracy and efficiency. For full hyperparameter details, refer to our code.

\begin{table}[htb]
\centering
\caption{Sensitivity of TAWRMAC to $M$ and $r$ hyperparameters.}
\label{tab:sensitivity}
\begin{tabular}{l |l c}
\hline
\textbf{Dataset} & \textbf{Configuration} & \textbf{AP} \\
\hline
\multirow{3}{*}{Enron} 
& $M=1$, $r=8$ & \textbf{77.28} \\
& $M=10$, $r=8$ & 73.49 \\
& $M=1$, $r=32$ & 73.82 \\
\hline
\multirow{3}{*}{UCI}
& $M=10$, $r=4$ & \textbf{82.85} \\
& $M=1$, $r=4$ & 82.03 \\
& $M=10$, $r=32$ & 78.71 \\
\hline
\multirow{3}{*}{Wikipedia}
& $M=1$, $r=128$ & \textbf{86.12} \\
& $M=10$, $r=128$ & 83.32 \\
& $M=1$, $r=32$ & 84.83 \\
\hline
\end{tabular}
\vspace{-5mm}
\end{table}

\subsection{Ablation Study}
\label{sec:ablation}
Figure~\ref{fig:ablation_study} presents the results of the ablation study, evaluating the contributions of each component of TAWRMAC on four datasets: MOOC, UN Vote, UCI, and Wikipedia. Variants without key modules (\textbf{w/o MAE}, \textbf{w/o NCE}, \textbf{w/o TAWR}, \textbf{w/o Restart}) exhibit noticeable performance drops, highlighting their significance. The MAE module proves to be the most critical, as its removal results in the largest performance degradation across most datasets. The NCE module is particularly important in UCI and UN Vote, where excluding it significantly reduces performance, suggesting that contextual neighborhood signals enhance predictive capacity. We also include a variant with Temporal Anonymous Walks but without the restart mechanism (\textbf{w/o Restart}) to demonstrate the importance of restart in distinguishing nodes with repetitive interactions from those exploring new connections. For instance, in UN Vote, removing restart lowers AP from 0.6671 to 0.628. TAWRMAC, in its full configuration, consistently achieves the highest AP across all datasets, showcasing its robustness and effectiveness. The error bars highlight TAWRMAC’s stability compared to its modified counterparts.






\section{Conclusion}
\label{sec:conclusion}
In this paper, we presented TAWRMAC, a novel framework for continuous-time dynamic graph representation learning. By integrating Memory-Augmented Embedding,  Neighbor Co-occurrence Embedding, and Temporal Anonymous Walks with Restart, TAWRMAC addressed key challenges in dynamic graph learning. Extensive experiments on multiple benchmark datasets demonstrated the superiority of TAWRMAC over existing methods in both dynamic link prediction and dynamic node classification tasks under various negative sampling strategies.

Despite its promising results, TAWRMAC has limitations in scalability, primarily due to the current implementation of Temporal Anonymous Walks with Restart. Currently, the model repeatedly samples a set of TAWRs from scratch for each new temporal link, leading to computational redundancy. A more efficient approach could involve maintaining a dynamic list of sampled TAWRs for each node, updated incrementally as new temporal links are processed. This would enhance computational efficiency.

Overall, TAWRMAC advances the state of the art in dynamic graph representation learning by providing a robust, generalizable, and context-aware approach to embedding evolving network structures. Future work will focus on further improving scalability, and exploring additional applications.

\clearpage
\bibliographystyle{ACM-Reference-Format}


\section{Additional Experimental Results}
\label{sec:exp_results_appendix}

Table~\ref{tab:ind_auc_comparison} in shows that TAWRMAC achieves the best average rank for AUC-ROC in inductive dynamic link prediction across all negative sampling strategies: 1.92 for random, 2.67 for historical, and 2.58 for inductive sampling. Similarly, Table~\ref{tab:trans_auc_comparison} presents AUC-ROC results for transductive dynamic link prediction, in which TAWRMAC continues to lead with the best average ranks of 1.5, 1.83, and 2.67 for random, historical, and inductive sampling, respectively.
\begin{table*}[htb]
\centering
\caption{AUC-ROC results for inductive dynamic link prediction.}
\label{tab:ind_auc_comparison}
\setlength{\tabcolsep}{2pt}
\scalebox{0.9}{
\begin{tabular}{p{14pt}|c|c|c|c|c|c|c|c|c|c}
\hline
\textbf{NSS}       & \textbf{Datasets}  & \textbf{JODIE} & \textbf{DyRep} & \textbf{TGAT} & \textbf{TGN} & \textbf{CAWN} & \textbf{TCL} & \textbf{GraphMixer} & \textbf{DyGFormer} & \textbf{TAWRMAC} \\
\hline
\multirow{13}{*}{\rotatebox{-90}{Random}} 
& Wikipedia  & 94.33 $\pm$ 0.27 & 91.49 $\pm$ 0.45 & 95.90 $\pm$ 0.09 & 97.72 $\pm$ 0.03 & 98.03 $\pm$ 0.04 & 95.57 $\pm$ 0.20 & 96.30 $\pm$ 0.04 & \underline{98.48 $\pm$ 0.03} & \textbf{98.91 $\pm$ 0.01} \\
& Reddit     & 96.52 $\pm$ 0.13 & 96.05 $\pm$ 0.12 & 96.98 $\pm$ 0.04 & 97.39 $\pm$ 0.07 & 98.42 $\pm$ 0.02 & 93.80 $\pm$ 0.07 & 94.97 $\pm$ 0.05 & \underline{98.71 $\pm$ 0.01} & \textbf{98.89 $\pm$ 0.01} \\
& MOOC       & 83.16 $\pm$ 1.30 & 84.03 $\pm$ 0.49 & 86.84 $\pm$ 0.17 & \underline{91.24 $\pm$ 0.99} & 81.86 $\pm$ 0.25 & 81.43 $\pm$ 0.19 & 82.77 $\pm$ 0.24 & 87.62 $\pm$ 0.51 & \textbf{92.13 $\pm$ 0.87} \\
& LastFM     & 81.13 $\pm$ 3.39 & 82.24 $\pm$ 1.51 & 76.99 $\pm$ 0.29 & 82.61 $\pm$ 3.15 & 87.82 $\pm$ 0.12 & 70.84 $\pm$ 0.85 & 80.37 $\pm$ 0.18 & \textbf{94.08 $\pm$ 0.08} & \underline{92.22 $\pm$ 0.07} \\
& Enron      & 81.96 $\pm$ 1.34 & 76.34 $\pm$ 4.20 & 64.63 $\pm$ 1.74 & 78.83 $\pm$ 1.11 & 87.02 $\pm$ 0.50 & 72.33 $\pm$ 0.99 & 76.51 $\pm$ 0.71 & \textbf{90.69 $\pm$ 0.26} & \underline{89.79 $\pm$ 0.35} \\
& UCI        & 78.80 $\pm$ 0.94 & 58.08 $\pm$ 1.81 & 77.64 $\pm$ 0.38 & 86.68 $\pm$ 2.29 & 90.40 $\pm$ 0.11 & 84.49 $\pm$ 1.82 & 89.30 $\pm$ 0.57 & \underline{92.63 $\pm$ 0.13} & \textbf{93.37 $\pm$ 0.28} \\
& Flights    & 95.21 $\pm$ 0.32 & 93.56 $\pm$ 0.70 & 88.64 $\pm$ 0.35 & 95.92 $\pm$ 0.43 & 96.86 $\pm$ 0.02 & 82.48 $\pm$ 0.01 & 82.27 $\pm$ 0.06 & \textbf{97.80 $\pm$ 0.02} & \underline{97.52 $\pm$ 0.07} \\
& Can. Parl. & 53.81 $\pm$ 1.14 & 55.27 $\pm$ 0.49 & 56.51 $\pm$ 0.75 & 55.86 $\pm$ 0.75 & \underline{58.83 $\pm$ 1.13} & 55.83 $\pm$ 1.07 & 58.32 $\pm$ 1.08 & \textbf{89.33 $\pm$ 0.48} & 57.15 $\pm$ 0.77 \\
& US Legis.  & 58.12 $\pm$ 2.35 & \underline{61.07 $\pm$ 0.56} & 48.27 $\pm$ 3.50 & \textbf{62.38 $\pm$ 0.48} & 51.49 $\pm$ 1.13 & 50.43 $\pm$ 1.48 & 47.20 $\pm$ 0.89 & 53.21 $\pm$ 3.04 & 59.78 $\pm$ 1.39 \\
& UN Trade   & 62.28 $\pm$ 0.50 & 58.82 $\pm$ 0.98 & 62.72 $\pm$ 0.12 & 59.99 $\pm$ 3.50 & \underline{67.05 $\pm$ 0.21} & 63.76 $\pm$ 0.07 & 63.48 $\pm$ 0.37 & \textbf{67.25 $\pm$ 1.05} & 66.92 $\pm$ 1.12 \\
& UN Vote    & 58.13 $\pm$ 1.43 & 55.13 $\pm$ 3.46 & 51.83 $\pm$ 1.35 & \textbf{61.23 $\pm$ 2.71} & 48.34 $\pm$ 0.76 & 50.51 $\pm$ 1.05 & 50.04 $\pm$ 0.86 & 56.73 $\pm$ 0.69 & \underline{60.04 $\pm$ 2.53} \\
& Contact    & 95.37 $\pm$ 0.92 & 91.89 $\pm$ 0.38 & 96.53 $\pm$ 0.10 & 94.84 $\pm$ 0.75 & 89.07 $\pm$ 0.34 & 93.05 $\pm$ 0.09 & 92.83 $\pm$ 0.05 & \underline{98.30 $\pm$ 0.02} & \textbf{98.49 $\pm$ 0.06} \\
\cline{2-11}
& Avg. Rank  & 5.75  & 6.67  & 6.25  & 4.08  & 4.50  & 7.25  & 6.50  & \underline{2.08}  & \textbf{1.92} \\

\Hline
\multirow{13}{*}{\rotatebox{-90}{Historical}} 
& Wikipedia  & 61.86 $\pm$ 0.53 & 57.54 $\pm$ 1.09 & 78.38 $\pm$ 0.20 & 75.75 $\pm$ 0.29 & 62.04 $\pm$ 0.65 & \underline{79.79 $\pm$ 0.96} & \textbf{82.87 $\pm$ 0.21} & 68.33 $\pm$ 2.82 & 77 $\pm$ 0.12 \\
& Reddit     & 61.69 $\pm$ 0.39 & 60.45 $\pm$ 0.37 & 64.43 $\pm$ 0.27 & 64.55 $\pm$ 0.50 & \underline{64.94 $\pm$ 0.21} & 61.43 $\pm$ 0.26 & 64.27 $\pm$ 0.13 & 64.81 $\pm$ 0.25 & \textbf{65.09 $\pm$ 0.46} \\
& MOOC       & 64.48 $\pm$ 1.64 & 64.23 $\pm$ 1.29 & 74.08 $\pm$ 0.27 & 77.69 $\pm$ 3.55 & 71.68 $\pm$ 0.94 & 69.82 $\pm$ 0.32 & 72.53 $\pm$ 0.84 & \textbf{80.77 $\pm$ 0.63} & \underline{77.73 $\pm$ 5.94} \\
& LastFM     & 68.44 $\pm$ 3.26 & 68.79 $\pm$ 1.08 & 69.89 $\pm$ 0.28 & 66.99 $\pm$ 5.62 & 67.69 $\pm$ 0.24 & 55.88 $\pm$ 1.85 & 70.07 $\pm$ 0.20 & \textbf{70.73 $\pm$ 0.37} & \underline{70.4 $\pm$ 0.16} \\
& Enron      & 65.32 $\pm$ 3.57 & 61.50 $\pm$ 2.50 & 57.84 $\pm$ 2.18 & 62.68 $\pm$ 1.09 & 62.25 $\pm$ 0.40 & 64.06 $\pm$ 1.02 & \underline{68.20 $\pm$ 0.96} & 65.78 $\pm$ 0.42 & \textbf{74.42 $\pm$ 0.35} \\
& UCI        & 60.24 $\pm$ 1.94 & 51.25 $\pm$ 2.37 & 62.32 $\pm$ 1.18 & 62.69 $\pm$ 0.90 & 56.39 $\pm$ 0.10 & 70.46 $\pm$ 1.94 & \underline{75.98 $\pm$ 0.84} & 65.55 $\pm$ 1.01 & \textbf{76.37 $\pm$ 0.17} \\
& Flights    & 60.72 $\pm$ 1.29 & 61.99 $\pm$ 1.39 & \underline{63.38 $\pm$ 0.26} & 59.66 $\pm$ 1.04 & 56.58 $\pm$ 0.44 & \textbf{63.48 $\pm$ 0.23} & 63.30 $\pm$ 0.19 & 56.05 $\pm$ 0.21 & 53.19 $\pm$ 0.55 \\
& Can. Parl. & 51.62 $\pm$ 1.00 & 52.38 $\pm$ 0.46 & 58.30 $\pm$ 0.61 & 55.64 $\pm$ 0.54 & \underline{60.11 $\pm$ 0.48} & 57.30 $\pm$ 1.03 & 56.68 $\pm$ 1.20 & \textbf{88.68 $\pm$ 0.74} & 57.9 $\pm$ 1.27 \\
& US Legis.  & 58.12 $\pm$ 2.94 & \textbf{67.94 $\pm$ 0.98} & 49.99 $\pm$ 4.88 & \underline{64.87 $\pm$ 1.65} & 54.41 $\pm$ 1.31 & 52.12 $\pm$ 2.13 & 49.28 $\pm$ 0.86 & 56.57 $\pm$ 3.22 & 58.83 $\pm$ 2.77 \\
& UN Trade   & 58.73 $\pm$ 1.19 & 57.90 $\pm$ 1.33 & 59.74 $\pm$ 0.59 & 55.61 $\pm$ 3.54 & 60.95 $\pm$ 0.80 & \underline{61.12 $\pm$ 0.97} & 59.88 $\pm$ 1.17 & 58.46 $\pm$ 1.65 & \textbf{61.33 $\pm$ 2.94} \\
& UN Vote    & 65.16 $\pm$ 1.28 & 63.98 $\pm$ 2.12 & 51.73 $\pm$ 4.12 & \underline{68.59 $\pm$ 3.11} & 48.01 $\pm$ 1.77 & 54.66 $\pm$ 2.11 & 45.49 $\pm$ 0.42 & 53.85 $\pm$ 2.02 & \textbf{69.76 $\pm$ 0.86} \\
& Contact    & 90.80 $\pm$ 1.18 & 88.88 $\pm$ 0.68 & \underline{93.76 $\pm$ 0.41} & 88.84 $\pm$ 1.39 & 74.79 $\pm$ 0.37 & 90.37 $\pm$ 0.16 & 90.04 $\pm$ 0.29 & \textbf{94.14 $\pm$ 0.26} & 93.62 $\pm$ 0.23 \\
\cline{2-11}
& Avg. Rank  & 5.92  & 6.75  & 4.83  & 5.42  & 6     & 4.92  & 4.67  & \underline{3.83}  & \textbf{2.67} \\

\Hline
\multirow{13}{*}{\rotatebox{-90}{Inductive}} 
& Wikipedia  & 61.87 $\pm$ 0.53 & 57.54 $\pm$ 1.09 & 78.38 $\pm$ 0.20 & 75.76 $\pm$ 0.29 & 62.02 $\pm$ 0.65 & 79.79 $\pm$ 0.96 & \textbf{82.88 $\pm$ 0.21} & 68.33 $\pm$ 2.82 & \underline{81.33 $\pm$ 1.61} \\
& Reddit     & 61.69 $\pm$ 0.39 & 60.44 $\pm$ 0.37 & 64.39 $\pm$ 0.27 & 64.55 $\pm$ 0.50 & \textbf{64.91 $\pm$ 0.21} & 61.36 $\pm$ 0.26 & 64.27 $\pm$ 0.13 & \underline{64.85 $\pm$ 0.25} & 64.8 $\pm$ 0.77 \\
& MOOC       & 64.48 $\pm$ 1.64 & 64.22 $\pm$ 1.29 & 74.07 $\pm$ 0.27 & 77.68 $\pm$ 3.55 & 71.69 $\pm$ 0.94 & 69.83 $\pm$ 0.32 & 72.52 $\pm$ 0.84 & \textbf{80.77 $\pm$ 0.63} & \underline{80.55 $\pm$ 2.63} \\
& LastFM     & 68.44 $\pm$ 3.26 & 68.79 $\pm$ 1.08 & 69.89 $\pm$ 0.28 & 66.99 $\pm$ 5.61 & 67.68 $\pm$ 0.24 & 55.88 $\pm$ 1.85 & 70.07 $\pm$ 0.20 & \underline{70.73 $\pm$ 0.37} & \textbf{72.38 $\pm$ 0.41} \\
& Enron      & 65.32 $\pm$ 3.57 & 61.50 $\pm$ 2.50 & 57.83 $\pm$ 2.18 & 62.68 $\pm$ 1.09 & 62.27 $\pm$ 0.40 & 64.05 $\pm$ 1.02 & \underline{68.19 $\pm$ 0.83} & 65.79 $\pm$ 0.42 & \textbf{74.63 $\pm$ 0.43} \\
& UCI        & 60.27 $\pm$ 1.94 & 51.26 $\pm$ 2.40 & 62.29 $\pm$ 1.17 & 62.66 $\pm$ 0.91 & 56.39 $\pm$ 0.11 & 70.42 $\pm$ 1.93 & \underline{75.97 $\pm$ 0.85} & 65.58 $\pm$ 1.00 & \textbf{76.42 $\pm$ 0.82} \\

& Flights    & 60.72 $\pm$ 1.29 & 61.99 $\pm$ 1.39 & \underline{63.4 $\pm$ 0.26} & 59.66 $\pm$ 1.05 & 56.58 $\pm$ 0.44 & \textbf{63.49 $\pm$ 0.23} & 63.32 $\pm$ 0.19 & 56.05 $\pm$ 0.22 & 53.38 $\pm$ 1.31 \\
& Can. Parl. & 51.61 $\pm$ 0.98 & 52.35 $\pm$ 0.52 & 58.15 $\pm$ 0.62 & 55.43 $\pm$ 0.42 & \underline{60.01 $\pm$ 0.47} & 56.88 $\pm$ 0.93 & 56.63 $\pm$ 1.09 & \textbf{88.51 $\pm$ 0.73} & 57.87 $\pm$ 1.27 \\
& US Legis.  & 58.12 $\pm$ 2.94 & \textbf{67.94 $\pm$ 0.98} & 49.99 $\pm$ 4.88 & \underline{64.87 $\pm$ 1.65} & 54.41 $\pm$ 1.31 & 52.12 $\pm$ 2.13 & 49.28 $\pm$ 0.86 & 56.57 $\pm$ 3.22 & 58.87 $\pm$ 3.48 \\
& UN Trade   & 58.71 $\pm$ 1.20 & 57.87 $\pm$ 1.36 & 59.98 $\pm$ 0.59 & 55.62 $\pm$ 3.59 & \underline{60.88 $\pm$ 0.79} & \textbf{61.01 $\pm$ 0.93} & 59.71 $\pm$ 1.17 & 57.28 $\pm$ 3.06 & 60.79 $\pm$ 1.7 \\
& UN Vote    & 65.29 $\pm$ 1.30 & 64.10 $\pm$ 2.10 & 51.78 $\pm$ 4.14 & \underline{68.58 $\pm$ 3.08} & 48.04 $\pm$ 1.76 & 54.65 $\pm$ 2.20 & 45.57 $\pm$ 0.41 & 53.87 $\pm$ 2.01 & \textbf{71.25 $\pm$ 0.86} \\
& Contact    & 90.8  $\pm$ 1.18 & 88.87 $\pm$ 0.67 & \underline{93.76 $\pm$ 0.40} & 88.85 $\pm$ 1.39 & 74.79 $\pm$ 0.38 & 90.37 $\pm$ 0.16 & 90.04 $\pm$ 0.29 & \textbf{94.14 $\pm$ 0.26} & 94.09 $\pm$ 0.23 \\
\cline{2-11}
& Avg. Rank  & 5.92  & 6.67  & 4.92  & 5.42  & 5.83  & 4.92  & 4.75  & \underline{4}  & \textbf{2.58} \\
\hline
\end{tabular}
}
\vspace{-2mm}
\end{table*}
\begin{table*}[htb]

\centering
\caption{AUC-ROC results for transductive dynamic link prediction.}
\label{tab:trans_auc_comparison}
\setlength{\tabcolsep}{2pt}
\scalebox{0.9}{
\begin{tabular}{p{14pt}|c|c|c|c|c|c|c|c|c|c|c}
\hline
\textbf{NSS}       & \textbf{Datasets} & \textbf{JODIE} & \textbf{DyRep} & \textbf{TGAT} & \textbf{TGN} & \textbf{CAWN} & \textbf{EdgeBank} & \textbf{TCL} & \textbf{GraphMixer} & \textbf{DyGFormer} & \textbf{TAWRMAC} \\ 
\hline

\multirow{13}{*}{\rotatebox{-90}{Random}} 
& Wikipedia  & 96.33 $\pm$ 0.07 & 94.37 $\pm$ 0.09 & 96.67 $\pm$ 0.07 & 98.37 $\pm$ 0.07 & 98.54 $\pm$ 0.04 & 90.78 $\pm$ 0.00 & 95.84 $\pm$ 0.18 & 96.92 $\pm$ 0.03 & \underline{98.91 $\pm$ 0.02} & \textbf{99.3 $\pm$ 0.01} \\
& Reddit     & 98.31 $\pm$ 0.05 & 98.17 $\pm$ 0.05 & 98.47 $\pm$ 0.02 & 98.6 $\pm$ 0.06 & 99.01 $\pm$ 0.01 & 95.37 $\pm$ 0.00 & 97.42 $\pm$ 0.02 & 97.17 $\pm$ 0.02 & \underline{99.15 $\pm$ 0.01} & \textbf{99.27 $\pm$ 0.01} \\
& MOOC       & 83.81 $\pm$ 2.09 & 85.03 $\pm$ 0.58 & 87.11 $\pm$ 0.19 & \underline{91.21 $\pm$ 1.15} & 80.38 $\pm$ 0.26 & 60.86 $\pm$ 0.00 & 83.12 $\pm$ 0.18 & 84.01 $\pm$ 0.17 & 87.91 $\pm$ 0.58 & \textbf{92.57 $\pm$ 0.6} \\
& LastFM     & 70.49 $\pm$ 1.66 & 71.16 $\pm$ 1.89 & 71.59 $\pm$ 0.18 & 78.47 $\pm$ 2.94 & 85.92 $\pm$ 0.10 & 83.77 $\pm$ 0.00 & 64.06 $\pm$ 1.16 & 73.53 $\pm$ 0.12 & \textbf{93.05 $\pm$ 0.10} & \underline{90.17 $\pm$ 0.14} \\
& Enron      & 87.96 $\pm$ 0.52 & 84.89 $\pm$ 3.00 & 68.89 $\pm$ 1.10 & 88.32 $\pm$ 0.99 & 90.45 $\pm$ 0.14 & 87.05 $\pm$ 0.00 & 75.74 $\pm$ 0.72 & 84.38 $\pm$ 0.21 & \underline{93.33 $\pm$ 0.13} & \textbf{93.97 $\pm$ 0.48} \\
& UCI        & 90.44 $\pm$ 0.49 & 68.77 $\pm$ 2.34 & 78.53 $\pm$ 0.74 & 92.03 $\pm$ 1.13 & 93.87 $\pm$ 0.08 & 77.30 $\pm$ 0.00 & 87.82 $\pm$ 1.36 & 91.81 $\pm$ 0.67 & \underline{94.49 $\pm$ 0.26} & \textbf{95.4 $\pm$ 0.22}  \\
& Flights    & 96.21 $\pm$ 1.42 & 95.95 $\pm$ 0.62 & 94.13 $\pm$ 0.17 & 98.22 $\pm$ 0.13 & 98.45 $\pm$ 0.01 & 90.23 $\pm$ 0.00 & 91.21 $\pm$ 0.02 & 91.13 $\pm$ 0.01 & \textbf{98.93 $\pm$ 0.01} & \textbf{98.93 $\pm$ 0.82} \\
& Can. Parl. & 78.21 $\pm$ 0.23 & 73.35 $\pm$ 3.67 & 75.69 $\pm$ 0.78 & 76.99 $\pm$ 1.80 & 75.70 $\pm$ 3.27 & 64.14 $\pm$ 0.00 & 72.46 $\pm$ 3.23 & \underline{83.17 $\pm$ 0.53} & \textbf{97.76 $\pm$ 0.41} & 82.39 $\pm$ 1.4 \\
& US Legis.  & \underline{82.85 $\pm$ 1.07} & 82.28 $\pm$ 0.32 & 75.84 $\pm$ 1.99 & \textbf{83.34 $\pm$ 0.43} & 77.16 $\pm$ 0.39 & 62.57 $\pm$ 0.00 & 76.27 $\pm$ 0.63 & 76.96 $\pm$ 0.79 & 77.9 $\pm$ 0.58 & 80.23 $\pm$ 1.11 \\
& UN Trade   & 69.62 $\pm$ 0.44 & 67.44 $\pm$ 0.83 & 64.01 $\pm$ 0.12 & 69.1 $\pm$ 1.67 & 68.54 $\pm$ 0.18 & 66.75 $\pm$ 0.00 & 64.72 $\pm$ 0.05 & 65.52 $\pm$ 0.51 & \underline{70.2 $\pm$ 1.44}  & \textbf{72.28 $\pm$ 1.4} \\
& UN Vote    & 68.53 $\pm$ 0.95 & 67.18 $\pm$ 1.04 & 52.83 $\pm$ 1.12 & \underline{69.71 $\pm$ 2.65} & 53.09 $\pm$ 0.22 & 62.97 $\pm$ 0.00 & 51.88 $\pm$ 0.36 & 52.46 $\pm$ 0.27 & 57.12 $\pm$ 0.62 & \textbf{70.7 $\pm$ 1.35} \\
& Contact    & 96.66 $\pm$ 0.89 & 96.48 $\pm$ 0.14 & 96.95 $\pm$ 0.08 & 97.54 $\pm$ 0.35 & 89.99 $\pm$ 0.34 & 94.34 $\pm$ 0.00 & 94.15 $\pm$ 0.09 & 93.94 $\pm$ 0.02 & \underline{98.53 $\pm$ 0.01} & \textbf{99.06 $\pm$ 0.02} \\\cline{2-12}
& Avg. Rank  & 5.17  & 6.58  & 7.08  & 3.5   & 5.08  & 8.17  & 8.5   & 6.92  & \underline{2.42}  & \textbf{1.5}   \\

\Hline

\multirow{13}{*}{\rotatebox{-90}{Historical}} 
& Wikipedia  & 80.77 $\pm$ 0.73 & 77.74 $\pm$ 0.33 & 82.87 $\pm$ 0.22 & 82.74 $\pm$ 0.32 & 67.84 $\pm$ 0.64 & 77.27 $\pm$ 0.00 & \underline{85.76 $\pm$ 0.46} & \textbf{87.68 $\pm$ 0.17} & 78.80 $\pm$ 1.95 & 85.28 $\pm$ 0.4 \\
& Reddit     & 80.52 $\pm$ 0.32 & 80.15 $\pm$ 0.18 & 79.33 $\pm$ 0.16 & \underline{81.11 $\pm$ 0.19} & 80.27 $\pm$ 0.30 & 78.58 $\pm$ 0.00 & 76.49 $\pm$ 0.16 & 77.80 $\pm$ 0.12 & 80.54 $\pm$ 0.29 & \textbf{82.31 $\pm$ 0.33} \\
& MOOC       & 82.75 $\pm$ 0.83 & 81.06 $\pm$ 0.94 & 80.81 $\pm$ 0.67 & \underline{88.00 $\pm$ 1.80} & 71.57 $\pm$ 1.07 & 61.90 $\pm$ 0.00 & 72.09 $\pm$ 0.56 & 76.68 $\pm$ 1.40 & 87.04 $\pm$ 0.35 & \textbf{88.22 $\pm$ 4.37} \\
& LastFM     & 75.22 $\pm$ 2.36 & 74.65 $\pm$ 1.98 & 64.27 $\pm$ 0.26 & 77.97 $\pm$ 3.04 & 67.88 $\pm$ 0.24 & 78.09 $\pm$ 0.00 & 47.24 $\pm$ 3.13 & 64.21 $\pm$ 0.73 & \underline{78.78 $\pm$ 0.35} & \textbf{78.99 $\pm$ 0.61} \\
& Enron      & 75.39 $\pm$ 2.37 & 74.69 $\pm$ 3.55 & 61.85 $\pm$ 1.43 & 77.09 $\pm$ 2.22 & 65.10 $\pm$ 0.34 & \underline{79.59 $\pm$ 0.00} & 67.95 $\pm$ 0.88 & 75.27 $\pm$ 1.14 & 76.55 $\pm$ 0.52 & \textbf{79.73 $\pm$ 0.48} \\
& UCI        & \underline{78.64 $\pm$ 3.50} & 57.91 $\pm$ 3.12 & 58.89 $\pm$ 1.57 & 77.25 $\pm$ 2.68 & 57.86 $\pm$ 0.15 & 69.56 $\pm$ 0.00 & 72.25 $\pm$ 3.46 & 77.54 $\pm$ 2.02 & 76.97 $\pm$ 0.24 & \textbf{82.76 $\pm$ 0.31} \\
& Flights    & 68.97 $\pm$ 1.87 & 69.43 $\pm$ 0.90 & \underline{72.20 $\pm$ 0.16} & 68.39 $\pm$ 0.95 & 66.11 $\pm$ 0.71 & \textbf{74.64 $\pm$ 0.00} & 70.57 $\pm$ 0.18 & 70.37 $\pm$ 0.23 & 68.09 $\pm$ 0.43 & 69.27 $\pm$ 0.83 \\
& Can. Parl. & 62.44 $\pm$ 1.11 & 70.16 $\pm$ 1.70 & 70.86 $\pm$ 0.94 & 73.23 $\pm$ 3.08 & 72.06 $\pm$ 3.94 & 63.04 $\pm$ 0.00 & 69.95 $\pm$ 3.70 & 79.03 $\pm$ 1.01 & \textbf{97.61 $\pm$ 0.40} & \underline{79.99 $\pm$ 1.66} \\
& US Legis.  & 67.47 $\pm$ 6.40 & \underline{91.44 $\pm$ 1.18} & 73.47 $\pm$ 5.25 & 83.53 $\pm$ 4.53 & 78.62 $\pm$ 7.46 & 67.41 $\pm$ 0.00 & 83.97 $\pm$ 3.71 & 85.17 $\pm$ 0.70 & 90.77 $\pm$ 1.96 & \textbf{92.95 $\pm$ 2.51} \\
& UN Trade   & 68.92 $\pm$ 1.40 & 64.36 $\pm$ 1.40 & 60.37 $\pm$ 0.68 & 63.93 $\pm$ 5.41 & 63.09 $\pm$ 0.74 & \textbf{86.61 $\pm$ 0.00} & 61.43 $\pm$ 1.04 & 63.20 $\pm$ 1.54 & 73.86 $\pm$ 1.13 & \underline{74.44 $\pm$ 0.83} \\
& UN Vote    & 76.84 $\pm$ 1.01 & 74.72 $\pm$ 1.43 & 53.95 $\pm$ 3.15 & 73.40 $\pm$ 5.20 & 51.27 $\pm$ 0.33 & \textbf{89.62 $\pm$ 0.00} & 52.29 $\pm$ 2.39 & 52.61 $\pm$ 1.44 & 64.27 $\pm$ 1.78 & \underline{79.15 $\pm$ 2.51} \\
& Contact    & 96.35 $\pm$ 0.92 & 96.00 $\pm$ 0.23 & 95.39 $\pm$ 0.43 & 93.76 $\pm$ 1.29 & 83.06 $\pm$ 0.32 & 92.17 $\pm$ 0.00 & 93.34 $\pm$ 0.19 & 93.14 $\pm$ 0.34 & \underline{97.17 $\pm$ 0.05} & \textbf{97.35 $\pm$ 0.1} \\\cline{2-12}
& Avg. Rank  & 5.17  & 5.67  & 6.75  & 4.58  & 8.33  & 5.83  & 7.08  & 5.75  & \underline{4}     & \textbf{1.83} \\

\Hline

\multirow{13}{*}{\rotatebox{-90}{Inductive}} 
& Wikipedia  & 70.96 $\pm$ 0.78 & 67.36 $\pm$ 0.96 & 81.93 $\pm$ 0.22 & 80.97 $\pm$ 0.31 & 70.95 $\pm$ 0.95 & 81.73 $\pm$ 0.00 & 82.19 $\pm$ 0.48 & \underline{84.28 $\pm$ 0.30} & 75.09 $\pm$ 3.70 & \textbf{85.36 $\pm$ 1.67} \\
& Reddit     & 83.51 $\pm$ 0.15 & 82.90 $\pm$ 0.31 & \underline{87.13 $\pm$ 0.20} & 84.56 $\pm$ 0.24 & \textbf{88.04 $\pm$ 0.29} & 85.93 $\pm$ 0.00 & 84.67 $\pm$ 0.29 & 82.21 $\pm$ 0.13 & 86.23 $\pm$ 0.51 & 85.14 $\pm$ 0.35 \\
& MOOC       & 66.63 $\pm$ 2.30 & 63.26 $\pm$ 1.01 & 73.18 $\pm$ 0.33 & 77.44 $\pm$ 2.86 & 70.32 $\pm$ 1.43 & 48.18 $\pm$ 0.00 & 70.36 $\pm$ 0.37 & 72.45 $\pm$ 0.72 & \underline{80.76 $\pm$ 0.76} & \textbf{82.56 $\pm$ 2.26} \\
& LastFM     & 61.32 $\pm$ 3.49 & 62.15 $\pm$ 2.12 & 63.99 $\pm$ 0.21 & 65.46 $\pm$ 4.27 & 67.92 $\pm$ 0.44 & \textbf{77.37 $\pm$ 0.00} & 46.93 $\pm$ 2.59 & 60.22 $\pm$ 0.32 & \underline{69.25 $\pm$ 0.36} & 67.83 $\pm$ 1.17 \\
& Enron      & 70.92 $\pm$ 1.05 & 68.73 $\pm$ 1.34 & 60.45 $\pm$ 2.12 & 71.34 $\pm$ 2.46 & \underline{75.17 $\pm$ 0.50} & 75.00 $\pm$ 0.00 & 67.64 $\pm$ 0.86 & 71.53 $\pm$ 0.85 & 74.07 $\pm$ 0.64 & \textbf{77.33 $\pm$ 0.46} \\
& UCI        & 64.14 $\pm$ 1.26 & 54.25 $\pm$ 2.01 & 60.80 $\pm$ 1.01 & 64.11 $\pm$ 1.04 & 58.06 $\pm$ 0.26 & 58.03 $\pm$ 0.00 & 70.05 $\pm$ 1.86 & \underline{74.59 $\pm$ 0.74} & 65.96 $\pm$ 1.18 & \textbf{75.13 $\pm$ 1.03} \\

& Flights    & 69.99 $\pm$ 3.10 & 71.13 $\pm$ 1.55 & \underline{73.47 $\pm$ 0.18} & 71.63 $\pm$ 1.72 & 69.70 $\pm$ 0.75 & \textbf{81.10 $\pm$ 0.00} & 72.54 $\pm$ 0.19 & 72.21 $\pm$ 0.21 & 69.53 $\pm$ 1.17 & 63.82 $\pm$ 1.91 \\
& Can. Parl. & 52.88 $\pm$ 0.80 & 63.53 $\pm$ 0.65 & 72.47 $\pm$ 1.18 & 69.57 $\pm$ 2.81 & 72.93 $\pm$ 1.78 & 61.41 $\pm$ 0.00 & 69.47 $\pm$ 2.12 & 70.52 $\pm$ 0.94 & \textbf{96.70 $\pm$ 0.59} & \underline{73.96 $\pm$ 1.77} \\
& US Legis.  & 59.05 $\pm$ 5.52 & \underline{89.44 $\pm$ 0.71} & 71.62 $\pm$ 5.42 & 78.12 $\pm$ 4.46 & 76.45 $\pm$ 7.02 & 68.66 $\pm$ 0.00 & 82.54 $\pm$ 3.91 & 84.22 $\pm$ 0.91 & 87.96 $\pm$ 1.80 & \textbf{89.81 $\pm$ 2.45} \\
& UN Trade   & 66.82 $\pm$ 1.27 & 65.60 $\pm$ 1.28 & 66.13 $\pm$ 0.78 & 66.37 $\pm$ 5.39 & \underline{71.73 $\pm$ 0.74} & \textbf{74.20 $\pm$ 0.00} & 67.80 $\pm$ 1.21 & 66.53 $\pm$ 1.22 & 62.56 $\pm$ 1.51 & 67.60 $\pm$ 2.45 \\
& UN Vote    & \underline{73.73 $\pm$ 1.61} & 72.80 $\pm$ 2.16 & 53.04 $\pm$ 2.58 & 72.69 $\pm$ 3.72 & 52.75 $\pm$ 0.90 & 72.85 $\pm$ 0.00 & 52.02 $\pm$ 1.64 & 51.89 $\pm$ 0.74 & 53.37 $\pm$ 1.26 & \textbf{75.31 $\pm$ 2.45} \\
& Contact    & 94.47 $\pm$ 1.08 & 94.23 $\pm$ 0.18 & 94.10 $\pm$ 0.41 & 91.64 $\pm$ 1.72 & 87.68 $\pm$ 0.24 & 85.87 $\pm$ 0.00 & 91.23 $\pm$ 0.19 & 90.96 $\pm$ 0.27 & \underline{95.01 $\pm$ 0.15} & \textbf{95.87 $\pm$ 0.11} \\\cline{2-12}
& Avg. Rank  & 6.75   & 7.17  & 5.58  & 5.67  & 5.58  & 5.42  & 5.92  & 5.83  & \underline{4.42}  & \textbf{2.67} \\ 
\hline

\end{tabular}
}

\end{table*}

\end{document}